%% file: sample_paper.tex
\documentclass[twoside,letterpaper]{article}
\usepackage[preprint]{aistats2027}
\usepackage[utf8]{inputenc} 
\usepackage{url}            
\usepackage{booktabs}       
\usepackage{amsfonts}       
\usepackage{graphicx}
\usepackage{tikz}
\usetikzlibrary{positioning, shapes.geometric, arrows.meta, fit}
\usepackage{algorithm}
\usepackage[noend]{algpseudocode}  
\usepackage[round,authoryear]{natbib} 
\usepackage{nicefrac}       
\usepackage[normalem]{ulem}  

\usepackage{microtype}      
\usepackage{amsmath,amssymb,amsfonts}
\usepackage{amsthm}
\usepackage{multirow}
\usepackage{xcolor}         
\usepackage[caption=false]{subfig}
\algrenewcommand{\algorithmiccomment}[1]{\hfill\(\triangleright\)\,\parbox[t]{0.32\linewidth}{#1}}
\newtheorem{theorem}{Theorem}[section]
\newtheorem{lemma}[theorem]{Lemma}  
\newtheorem{remark}{Remark}

\usepackage[hidelinks]{hyperref}

\begin{document}
\runningtitle{Certified Robustness via Sparse SOCP Relaxations}
\runningauthor{Maleki, Sadig, Sidibomma, Fazlyab, Adibi, and Samavi}
\twocolumn[
\aistatstitle{Certified Robustness via Sparse Second-Order Cone Relaxations}
\aistatsauthor{
Mohammadreza Maleki \And Daniel Sadig \And Rushendra Sidibomma
}
\aistatsaddress{
\small Toronto Metropolitan University \And \small Toronto Metropolitan University \And
\small University of Minnesota Twin Cities
}
\aistatsauthor{
Mahyar Fazlyab \And Arman Adibi \And Reza Samavi
}
\aistatsaddress{
\small Johns Hopkins University \And \small Augusta University \And
\small Toronto Metropolitan University, Vector Institute
}
]
\begingroup
\setlength{\leftmargini}{0.25in}
\begin{abstract}
  Certified robustness is critical for deploying neural networks in safety-sensitive settings, but computing exact certificates typically requires solving nonconvex optimization problems. Practical verifiers use convex relaxations: 
  linear programming (LP) relaxations are scalable but often too conservative, whereas semidefinite programming (SDP) relaxations can be substantially tighter but are computationally expensive. We introduce a hierarchy of second-order cone programming (SOCP) relaxations that provide stronger certificates than LP while avoiding the cost of global positive-semidefinite constraints. Our construction combines convex-hull relaxations of ReLU activations, selective McCormick lifting over certified variable bounds, and second-order cone constraints induced by $2\times2$ principal minors. Adding these constraints progressively yields a monotone hierarchy of increasingly tight bounds with a controllable computation-tightness trade-off. 
  On fully connected and convolutional networks on MNIST and CIFAR-10, SOCP-based verification improves certified robust accuracy over LP-based verification and is competitive with \(\alpha,\beta\)-CROWN, while remaining orders of magnitude faster than SDP-based verification. These results position selective second-order constraints as a practical middle ground between scalable LP-based and tight SDP-based certification. Finally, we introduce class-optimized robust evaluation (CORE), which applies tighter relaxations only to target classes not certified by cheaper ones. On the fully connected MNIST benchmark, CORE variants match SDP-cert’s certified robust accuracy while reducing average verification time by up to \(86.30\%\). 
\end{abstract}
\endgroup

\input{sections/Introduction}
\input{sections/Methodology}
\input{sections/Experiment}
\input{sections/Conclusion}
\label{migration:main-end}

\bibliography{references}
\clearpage
\appendix
\onecolumn
\input{sections/Appendix}
\end{document}

%% file: sections/Introduction.tex
\section{Introduction}

Reliable deployment of neural networks in safety-sensitive settings requires guarantees that extend beyond accuracy on unperturbed inputs. Small adversarial perturbations can change a classifier's prediction while remaining visually imperceptible~\citep{madry2017towards, sadig2026cear}. This vulnerability motivates \emph{certified robustness}, which provides a mathematical guarantee that the classifier assigns the correct label to every input in a prescribed perturbation set~\citep{wong2018provable,raghunathan2018semidefinite}. 

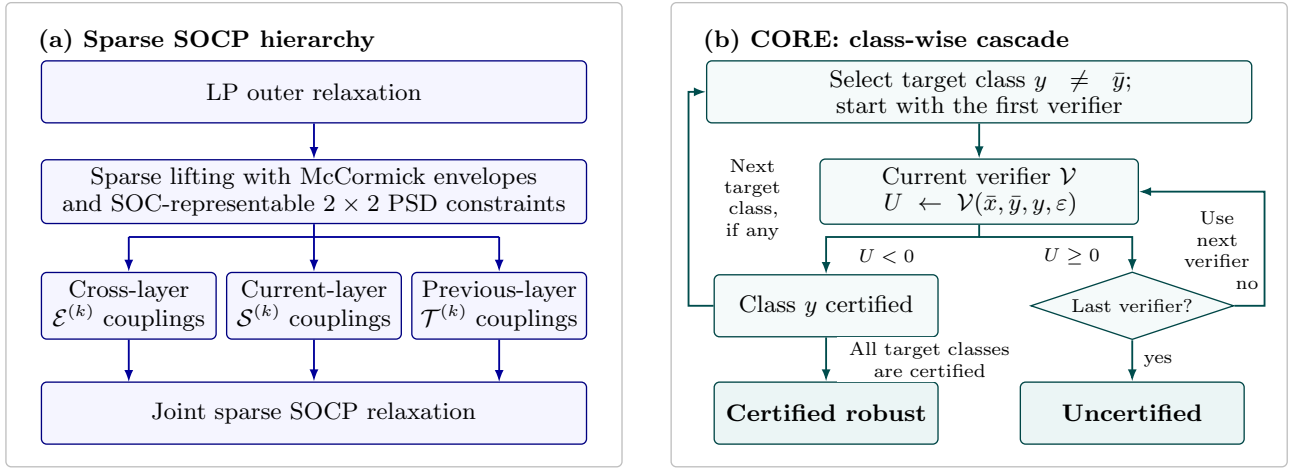
\begin{figure*}[t]
\centering
\input{Figures/overview-review}
\caption{Overview of the proposed methods: (a) jointly selected cross-layer, current-layer, and previous-layer couplings at each hidden layer, with \(z^{(0)}=x\), and (b) CORE's class-wise cascade with early stopping.}
\label{fig:overview}
\end{figure*}

Exact certification can be formulated by maximizing each competing class's logit margin relative to the true class over the allowed perturbations. Computing these worst-case margins generally requires solving nonconvex optimization problems, which can make verification computationally hard~\citep{katz2017reluplex}. For ReLU networks, complete verifiers, including satisfiability modulo theories (SMT)~\citep{pulina2012challenging}, mixed-integer optimization~\citep{tjeng2019evaluating}, and branch-and-bound methods~\citep{wang2021beta}, can resolve a verification problem given sufficient computation, but their cost can be prohibitive. Incomplete verifiers instead bound these margins to certify that the prediction remains correct under every admissible perturbation. If these certification bounds cannot be established, the input remains uncertified. A robust input left uncertified represents a false negative in robustness certification.


In this paper, we introduce a hierarchy of sparse second-order cone programming (SOCP) relaxations with a controllable trade-off between bound tightness and computational cost, and a class-wise cascade to progressively  allocate more expensive verifiers to unresolved margins. Figure~\ref{fig:overview} illustrates both components.

\textbf{Sparse SOCP hierarchy (Fig.~\ref{fig:overview}a).}
Starting from the exact quadratically constrained quadratic program (QCQP) for ReLU-network verification, we combine ReLU convex-hull relaxations~\citep{wong2018provable} with sparse lifting of selected cross-layer, current-layer, and previous-layer products at each hidden layer. These couplings are incorporated jointly using McCormick envelopes over certified variable bounds~\citep{mccormick1976computability} and second-order cone (SOC) representations of \(2\times2\) positive-semidefinite (PSD) constraints~\citep{kim2001second}, avoiding a global PSD constraint. Increasing the number of selected couplings tightens the relaxation at the expense of greater computational cost. Experiments conducted on MNIST~\citep{LeCun_Burges_Cortes_2010} and CIFAR-10~\citep{Krizhevsky_2012} illustrate that SOCP-based verification improves certified robust accuracy over the linear programming (LP) verifier and is competitive with \(\alpha,\beta\)-CROWN under the same evaluated settings. On the fully connected MNIST benchmark, SOCP verifier is orders of magnitude faster than the semidefinite programming (SDP) verifier.

\textbf{Class-optimized robust evaluation (CORE) (Fig.~\ref{fig:overview}b).}
CORE allocates verification effort to individual class-wise margin problems. For each target class, it applies verifiers in sequence, moving to the next class as soon as robustness against the current class is certified. CORE certifies an input as robust once every target class is certified. If a class remains unresolved at the final stage, CORE stops and leaves the input uncertified. Therefore, CORE maintains at least the certified robust accuracy of the final verifier while avoiding unnecessary computation. On the fully connected MNIST benchmark, CORE variants match SDP-cert's certified robust accuracy while reducing average verification time by up to \(86.30\%\).
\paragraph{Related work.}
Convex relaxations balance certification tightness and computational cost. LP relaxations of ReLU networks~\citep{wong2018provable} support scalable dual formulations and bound propagation~\citep{wong2018scaling,zhang2018efficient}, with general cutting planes providing additional tightening~\citep{zhang2022general}. Building on this framework, \(\alpha,\beta\)-CROWN combines optimized linear bound propagation with branch-and-bound for complete verification~\citep{xu2020fast,wang2021beta}. However, relaxing neurons separately can fail to capture dependencies among their activations. SDP relaxations capture second-order interactions through lifted matrices and global PSD constraints~\citep{raghunathan2018semidefinite,vandenberghe1996semidefinite}. Quadratic-constraint formulations~\citep{fazlyab2020safety}, memory-efficient solvers~\citep{dathathri2020enabling}, and linear reformulations~\citep{lan2022tight} improve their practicality, although global PSD constraints remain costly. SDP-CROWN incorporates \(\ell_2\)-norm-based inter-neuron coupling into bound propagation~\citep{chiu2025sdp}. Cascading robustness verification (CRV)~\citep{maleki2026cascading} reduces verification cost by escalating only uncertified inputs to more expensive verifiers. These approaches motivate capturing useful second-order structure without a global PSD constraint and allocating tighter verification to unresolved class-wise margins.

In this paper, we first formulate the robustness verification problem and then develop the sparse SOCP relaxations and CORE (Sections~\ref{sec:methodology}--\ref{sec:core}). We then evaluate these methods experimentally (Section~\ref{sec:experiment}) before presenting the conclusion and limitations (Section~\ref{sec:conclusion}).

%% file: Figures/overview-review.tex
\begin{tikzpicture}[
  x=1cm,y=1cm,
  panel/.style={draw=black!25,fill=white,rounded corners=2pt,line width=0.5pt},
  stage/.style={draw=blue!50!black,fill=blue!4,rounded corners=2pt,
    line width=0.5pt,align=center,text=black,font=\footnotesize,
    inner sep=3pt,text width=7.0cm,minimum height=0.82cm},
  task/.style={draw=teal!55!black,fill=teal!4,rounded corners=2pt,
    line width=0.5pt,align=center,text=black,font=\footnotesize,
    inner sep=3pt,minimum height=0.82cm},
  decision/.style={draw=teal!55!black,fill=teal!4,diamond,aspect=3.0,minimum width=2.3cm,minimum height=0.82cm,
    line width=0.5pt,align=center,text=black,font=\footnotesize,inner sep=2pt},
  terminal/.style={task,minimum height=0.82cm,fill=teal!8,font=\footnotesize\bfseries},
  hierarchy/.style={-{Latex[length=1.6mm]},draw=blue!60!black,line width=0.7pt},
  flow/.style={-{Latex[length=1.6mm]},draw=teal!60!black,line width=0.7pt},
  edge/.style={font=\scriptsize,text=black,fill=white,inner sep=2pt,align=center},
  title/.style={font=\small\bfseries,text=black,anchor=north west}
]
\draw[panel] (0,0) rectangle (8.15,-6.14);
\draw[panel] (8.8,0) rectangle (16.95,-6.14);
\node[title] at (0.30,-0.22) {(a) Sparse SOCP hierarchy};
\node[title] at (9.10,-0.22) {(b) CORE: class-wise cascade};

\node[stage] (lp) at (4.075,-1.18)
  {LP outer relaxation};
\node[stage] (lift) at (4.075,-2.50)
  {Sparse lifting with McCormick envelopes\\
   and SOC-representable \(2\times2\) PSD constraints};
\node[stage,text width=2.10cm] (cross) at (1.625,-4.01)
  {Cross-layer\\\(\mathcal E^{(k)}\) couplings};
\node[stage,text width=2.10cm] (current) at (4.075,-4.01)
  {Current-layer\\\(\mathcal S^{(k)}\) couplings};
\node[stage,text width=2.10cm] (previous) at (6.525,-4.01)
  {Previous-layer\\\(\mathcal T^{(k)}\) couplings};
\node[stage] (joint) at (4.075,-5.43)
  {Joint sparse SOCP relaxation};
\draw[hierarchy] (lp.south) -- (lift.north);
\draw[draw=blue!60!black,line width=0.7pt] (lift.south) -- (4.075,-3.10)
  -- (1.625,-3.10);
\draw[draw=blue!60!black,line width=0.7pt] (4.075,-3.10) -- (6.525,-3.10);
\draw[hierarchy] (1.625,-3.10) -- (cross.north);
\draw[hierarchy] (4.075,-3.10) -- (current.north);
\draw[hierarchy] (6.525,-3.10) -- (previous.north);
\draw[hierarchy] (cross.south) -- (1.625,-5.02);
\draw[hierarchy] (current.south) -- (joint.north);
\draw[hierarchy] (previous.south) -- (6.525,-5.02);

\node[task,text width=7.0cm] (start) at (12.875,-1.18)
  {Select target class \(y\ne\bar y\); start with the first verifier};
\node[task,text width=4.0cm,minimum height=0.82cm] (eval) at (12.875,-2.50)
  {Current verifier \(\mathcal V\)\\
   \(U\gets\mathcal V(\bar x,\bar y,y,\varepsilon)\)};
\node[task,text width=2.75cm,minimum height=0.82cm] (cert) at (10.85,-4.01)
  {Class \(y\) certified};
\node[decision,font=\scriptsize] (more) at (14.90,-4.01)
  {Last verifier?};
\node[terminal,text width=2.75cm] (robust) at (10.85,-5.43)
  {Certified robust};
\node[terminal,text width=2.75cm] (unknown) at (14.90,-5.43)
  {Uncertified};

\draw[flow] (start.south) -- (eval.north);
\draw[flow] (eval.south) -- (12.875,-3.10) -- (10.85,-3.10) -- (cert.north);
\draw[flow] (12.875,-3.10) -- (14.90,-3.10) -- (more.north);
\node[edge] at (11.65,-3.35) {\(U<0\)};
\node[edge] at (14.10,-3.35) {\(U\ge0\)};
\draw[flow] (cert.south) -- (robust.north)
  node[edge,midway,right,xshift=4pt,text width=2.3cm] {All target classes\\are certified};
\draw[flow] (cert.west) -- (9.03,-4.01) -- (9.03,-1.18) -- (start.west);
\node[edge,anchor=west,text width=1.35cm] at (9.12,-2.60) {Next target\\class, if any};
\draw[flow] (more.east) -- (16.65,-4.01) -- (16.65,-2.50) -- (eval.east);
\node[edge,text width=1.0cm] at (16.02,-3.13) {Use next\\verifier};
\node[edge,anchor=south,fill=none] at (16.42,-3.86) {no};
\draw[flow] (more.south) -- (unknown.north)
  node[edge,midway,right,xshift=2pt] {yes};
\end{tikzpicture}

%% file: sections/Methodology.tex
\section{Robustness Verification}\label{sec:methodology}

Consider a neural network \( f: \mathcal{X} \rightarrow \mathbb{R}^L\) with one hidden layer consisting of \( m \) neurons. The input space is \( \mathcal{X} \subseteq \mathbb{R}^d \), where \( d \) is the input dimension, and the output space represents scores (or logits) across \( L \) classes. The network computes the logits as $f(x)=W_2\,\mathrm{ReLU}(W_1 x)$, where
$W_1 \in \mathbb{R}^{m \times d}$ and $W_2 \in \mathbb{R}^{L \times m}$
denote the input-to-hidden and hidden-to-output weight matrices, respectively. Let the hidden pre-activation and activation be
\(s \triangleq W_1x\) and \(z \triangleq \mathrm{ReLU}(s)\).  For simplicity, we omit bias terms, though they are included in our experiments. We use one hidden layer for ease of presentation; the framework extends layer-wise to multilayer ReLU networks (Appendix~\ref{sec:socp-multilayer}). 


Given an input \( x \in \mathcal{X} \), the network assigns label \( y \) by selecting the class with the highest score,
\begin{equation}
y \triangleq \underset{1 \leq i \leq L}{\arg\max} f(x)_i,
\end{equation}
where \( f(x)_i\) is the \(i^{th}\) element of \( f(x)\). To certify  robustness against input perturbations, we consider an \(\ell_\infty\)-ball centered on a correctly classified input \(\bar{x}\), defined as
\[
B_{\varepsilon}(\bar{x}) \triangleq \{x \mid \|x - \bar{x}\|_{\infty} \leq \varepsilon\}.\]
Let \(\bar y\) denote the true label of the clean input \(\bar x\). The worst-case margin \( f^\star_y(\bar{x}) \) quantifies the robustness of the model at input \(\bar{x}\) by maximizing the difference between the logit score of any target class \( y \neq \bar{y} \), \(f(x)_{y}\), and the true class \( \bar{y} \), \(f(x)_{\bar{y}} \), over all valid perturbations:
\begin{equation}\label{eq:verification-method}
f^\star_y(\bar{x}) \triangleq \max_{x \in B_{\varepsilon}(\bar{x})} \big(f(x)_{y} - f(x)_{\bar{y}}\big).
\end{equation}
An input \(\bar{x}\) is certifiably robust if, for all \( y \neq \bar{y} \), the worst-case margin satisfies \( f^\star_y(\bar{x}) < 0\), ensuring no perturbation within \( B_{\varepsilon}(\bar{x}) \) leads to misclassification. Conversely, \(\bar{x}\) is considered non-robust if there exists a \( y \neq \bar{y} \) such that \( f^\star_y(\bar{x}) \geq 0 \), indicating a potential adversarial example within the allowed perturbation level \(\varepsilon\).

For a fixed target class \(y\neq\bar{y}\), let
\(c\triangleq W_2^\top(e_y-e_{\bar{y}})\), where \(e_i\) is the \(i\)th standard basis vector in \(\mathbb{R}^L\).
Then
\(
f(x)_y-f(x)_{\bar{y}}=c^\top z.
\)
Thus, the class-wise verification problem maximizes a linear function of the hidden activations over all admissible inputs and their corresponding ReLU activations.


Solving~\eqref{eq:verification-method} is generally intractable due to the non-convexity introduced by ReLU activations. Although complete verification methods exist, they are often computationally prohibitive, with worst-case complexity that grows exponentially with network size~\citep{raghunathan2018semidefinite, huang2017adversarial}. Incomplete verifiers based on convex relaxations such as LP, SDP, and dual formulations derive an upper bound \( f^{u}_y(\bar{x})\) satisfying \( f^{u}_y(\bar{x}) \geq f^\star_y(\bar{x}) \), reducing robustness certification to verifying whether \( f^{u}_y(\bar{x}) < 0 \) for every target class \(y\ne\bar y\).  In contrast, \( f^{u}_y(\bar{x}) \geq 0 \) is inconclusive for an incomplete verifier; i.e., failure to verify an input does not necessarily imply that the input is non-robust.

\paragraph{Exact quadratic formulation.}
The perturbation set $B_{\varepsilon}(\bar{x})$ is the box \([l,u]\), where
\(l\triangleq\bar{x}-\varepsilon\mathbf{1}\) and
\(u\triangleq\bar{x}+\varepsilon\mathbf{1}\).
This box constraint can be equivalently expressed as the quadratic constraint,
\begin{equation}\label{eq:input}
    (x - l)\odot(x - u) \le 0 \;\;\Rightarrow\;\; x \odot x \le (l+u)\odot x - l\odot u,
\end{equation}
where \(\odot\) denotes element-wise multiplication and vector inequalities are interpreted componentwise. Furthermore, the ReLU relation \(z=\mathrm{ReLU}(s)\) can be equivalently encoded as \(z \ge 0\), \(z \ge s\), and \(z\odot(z-s)=0\).
These constraints ensure that \(z\) behaves as a ReLU activation, maintaining equivalence to the original function. Together with the input constraint in~\eqref{eq:input} and the affine
relation \(s=W_1x\), this gives the following exact QCQP for the
one-hidden-layer verification problem~\citep{raghunathan2018semidefinite}:
\begin{align}\label{eq:qcqp-relax-relu-2}
	\begin{split}
            f_{\mathrm{QCQP}}^\star (\bar{x}) & \;\triangleq\;\max_{x,z} \  c^\top z \\
            \text{s.t.}\quad
             & z \odot z = z \odot s, \quad s = W_1x,\quad z \ge 0,\\
            & x\odot x \le (l + u) \odot x - l \odot u, \quad z \ge s.
	\end{split}
\end{align}
Following \citet{raghunathan2018semidefinite}, we lift this QCQP by defining 
\(v^\top \triangleq \begin{bmatrix} 1 & x^\top & z^\top \end{bmatrix}\) and \(P \triangleq vv^\top.\) The rank-one constrained optimization problem in~\eqref{eq:qcqp-relax-relu-2} can be written in terms of the matrix \(P\) as follows.
\begin{align}\label{eq:qcqp-rank1}
\begin{split}
f_{\mathrm{Rank1}}^\star (\bar{x}) &\;\triangleq\; \  \max_{P\succeq 0}\ c^\top P[z] \\
\text{s.t.}\quad
& \mathrm{diag}\!\big(P[zz^\top]\big)
  = \mathrm{diag}\!\big(W_1P[xz^\top]\big), \\
& P[z] \ge 0,\quad P[z] \ge W_1 P[x],\\
& \mathrm{diag}\!\big(P[xx^\top]\big) \le (l + u) \odot P[x] - l \odot u, \\
& P[1] = 1,\quad \mathrm{rank}(P) = 1.
\end{split}
\end{align}
The rank constraint is the only non-convex constraint in
\eqref{eq:qcqp-rank1}. Removing the rank-one constraint yields an SDP relaxation whose optimal value provides an upper bound on the worst-case margin \(f^\star_y(\bar x)\). LP relaxations~\citep{wong2018provable} are efficient but can be loose, while SDP relaxations~\citep{raghunathan2018semidefinite} are typically tighter but significantly more expensive. These observations motivate an \textit{intermediate relaxation} that balances tightness and scalability. We develop this relaxation using \textit{SOCP} in the next section.

\section{SOCP Relaxation}\label{sec:pure-socp}
Starting from the lifted rank-one formulation~\eqref{eq:qcqp-rank1}, the canonical SDP relaxation drops the non-convex constraint $\mathrm{rank}(P)=1$ while retaining $P\succeq 0$, yielding strong certificates but scaling poorly as the network size or verification problem size increases~\citep{raghunathan2018semidefinite}. Our goal is to preserve the main benefit of SDP, namely, its ability to capture second-order interactions among lifted variables, while avoiding the cost of enforcing global positive semidefiniteness. We build SOCP outer relaxations by lifting selected quadratic and bilinear terms into auxiliary variables, enforcing their consistency over known bounds via convex envelopes, and replacing $P\succeq 0$ with SOC-representable surrogates that enforce positive semidefiniteness only on a carefully chosen sparse collection of low-dimensional principal submatrices.

\subsection{Sparse SOCP Relaxation}\label{sec:sparse-socp}
We derive a scalable second-order cone relaxation of the rank-one lifted
formulation in~\eqref{eq:qcqp-rank1} by combining
(i) a convex-hull relaxation of each ReLU, which removes the single-neuron
quadratic complementarity constraint;
(ii) sparse lifting of selected bilinear terms via McCormick envelopes,
which introduces auxiliary variables only for targeted products and bounds
them using the available lower and upper bounds on their factors; and
(iii) local PSD surrogates through selected \(2\times2\) principal minors,
which enforce second-order consistency among selected lifted variables while
remaining exactly SOCP-representable.

\begin{figure*}[t]
    \centering
    \includegraphics[width=0.70\linewidth]{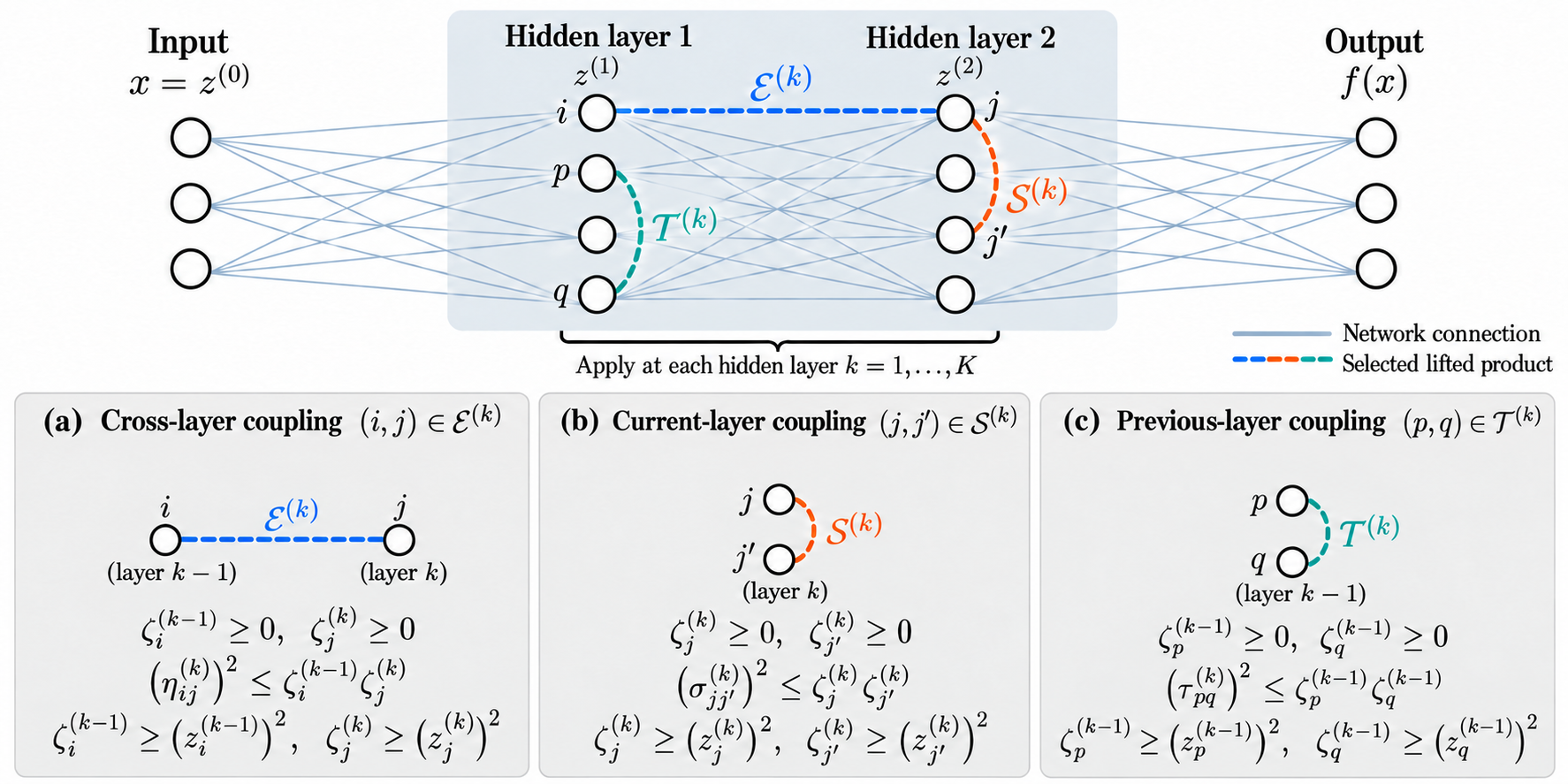}
    \caption{ Panels (a)--(c) illustrate sparse SOCP couplings by selecting
    cross-layer \(\mathcal E^{(k)}\), current-layer \(\mathcal S^{(k)}\),
    and previous-layer \(\mathcal T^{(k)}\) couplings and their
    SOC-representable \(2\times2\) PSD constraints.
    Square lifts are shared across incident products, which also satisfy
    McCormick envelopes.}
    \label{fig:SOCP}
\end{figure*}

Figure~\ref{fig:SOCP} shows three coupling types at hidden layer \(k\): cross-layer \(\mathcal E^{(k)}\) selects \(z_i^{(k-1)}z_j^{(k)}\), current-layer \(\mathcal S^{(k)}\) selects \(z_j^{(k)}z_{j'}^{(k)}\), and previous-layer \(\mathcal T^{(k)}\) selects \(z_p^{(k-1)}z_q^{(k-1)}\). Here \(z^{(k)}\) is the activation vector at layer \(k\), with \(z^{(0)}=x\). Each selected product satisfies McCormick envelopes and a local \(2\times2\) PSD constraint; square lifts are shared across incident products and adjacent layers. The derivation below uses \(k=1\), omitting layer superscripts; Section~\ref{rem:multilayer} and Appendix~\ref{sec:socp-multilayer} give the extension.

\paragraph{Base Relaxation.}
We replace the complementarity constraint \(z\odot(z-s)=0\) in
\eqref{eq:qcqp-relax-relu-2} with the convex-hull relaxation of each ReLU
over bounded pre-activations~\citep{wong2018provable}. Let
\(s=W_1x\), and suppose that bounds
\(\ell_{s,j}\le s_j\le u_{s,j}\) are available for each hidden unit \(j\). The corresponding activation upper bound is \(u_{z,j}=\max\{0,u_{s,j}\}\).
For an unstable neuron, i.e., \(\ell_{s,j}<0<u_{s,j}\), we impose
\begin{equation}\label{convex-hull}
\begin{gathered}
    z_j \ge 0,\quad
    z_j \ge s_j,\quad
    z_j \le \alpha_j(s_j-\ell_{s,j}),\\
    \alpha_j \triangleq
    \frac{u_{s,j}}{u_{s,j}-\ell_{s,j}} .
\end{gathered}
\end{equation}
For stable neurons, the relaxation is exact: if \(\ell_{s,j}\ge 0\), then
\(z_j=s_j\), while if \(u_{s,j}\le 0\), then \(z_j=0\).

We next introduce auxiliary diagonal moment variables $\{\xi_i\}_{i=1}^d$ to lift \(x_i^2\).
For the input box $x_i\in[\ell_{x,i},u_{x,i}]$, the quadratic implication $(x_i-\ell_{x,i})(x_i-u_{x,i})\le 0$ yields the affine upper bound
$x_i^2 \le (\ell_{x,i}+u_{x,i})x_i-\ell_{x,i}u_{x,i}$.
We lift this relation by replacing $x_i^2$ with $\xi_i$ and enforcing
\begin{equation}\label{eq:input-ss}
0 \le \xi_i \le (\ell_{x,i}+u_{x,i})x_i - \ell_{x,i}u_{x,i},\qquad i\in [d].
\end{equation}
To capture a subset of cross-layer couplings, which are input-activation products for \(k=1\), without incurring the full
\(\mathcal O(dm)\) lifting cost, we select a sparse edge set
\(\mathcal E\subseteq[d]\times[m]\). For each \((i,j)\in\mathcal E\), we
introduce an auxiliary variable \(\eta_{ij}\) for the bilinear product
\(x_i z_j\). Given bounds \(x_i\in[\ell_{x,i},u_{x,i}]\) and
\(z_j\in[0,u_{z,j}]\), we impose the McCormick envelope
\citep{mccormick1976computability}:
\begin{equation}\label{eq:mccormick-xz}
\begin{aligned}
\eta_{ij} &\ge \ell_{x,i}z_j, \\
\eta_{ij} &\ge u_{x,i}z_j + u_{z,j}x_i - u_{x,i}u_{z,j},\\
\eta_{ij} &\le u_{x,i}z_j,\\
\eta_{ij} &\le \ell_{x,i}z_j + u_{z,j}x_i - \ell_{x,i}u_{z,j},
\quad (i,j)\in\mathcal E .
\end{aligned}
\end{equation}
For any \(x_i\) and \(z_j\) within these bounds, the exact product
\(\eta_{ij}=x_i z_j\) satisfies all four inequalities. Thus,
\eqref{eq:mccormick-xz} contains the exact product relation but also permits
additional convexified points, giving a valid outer relaxation.

We then add a second-order consistency constraint so that the relaxed products remain tied to the variables from which they were lifted.
For each selected pair \((i,j)\in\mathcal E\), we introduce diagonal
surrogates \(\xi_i\) and \(\zeta_j\) for \(x_i^2\) and \(z_j^2\), and impose
the local PSD constraint
\begin{equation}\label{eq:soc-minors}
\begin{bmatrix}
\xi_i & \eta_{ij}\\
\eta_{ij} & \zeta_j
\end{bmatrix}
\succeq 0,
\quad
\xi_i \ge x_i^2,\quad
\zeta_j \ge z_j^2,
\quad (i,j)\in\mathcal E.
\end{equation}
Since a \(2\times2\) PSD constraint is equivalent to
\(\xi_i,\zeta_j\ge 0\) and \(\eta_{ij}^2\le \xi_i\zeta_j\), it admits a
rotated-SOC representation.

We then use the ReLU relation to define the activation-square surrogate from the selected lifted products, while bounding the contribution of the products that are not lifted.
In the exact lifted model, the ReLU complementarity condition implies
\begin{equation}
    z_j^2 = \sum_{i=1}^d (W_1)_{ji} \ x_i \ z_j.
\end{equation}
Since we introduce bilinear variables \(\eta_{ij}\) only for selected pairs
\((i,j)\in\mathcal E\), we represent these selected terms explicitly and
absorb the omitted terms into a residual:
\begin{equation}\label{eq:zeta-residual-main}
    \zeta_j
    =
    \sum_{i:(i,j)\in\mathcal E} (W_1)_{ji}\eta_{ij}
    + r_j,
    \qquad
    r_j \in [L^{(r)}_j,U^{(r)}_j].
\end{equation}
The residual \(r_j\) accounts for the terms with \((i,j)\notin\mathcal E\); its interval bounds are computed from the bounds on \(x_i\) and \(z_j\), as described in Appendix~\ref{app:residual-bounds}.

\paragraph{Additional Sparse Couplings.}\label{sec:additional-couplings}
The base relaxation can be tightened by adding selected current-layer and previous-layer couplings.
We first lift selected current-layer activation products. Let \(\mathcal S\subseteq[m]\times[m]\) with \(j<j'\), and introduce a surrogate \(\sigma_{jj'}\) for \(z_jz_{j'}\), for each \((j,j')\in\mathcal S\). We impose a convex outer relaxation over
$z_j\in[0,u_{z,j}]$ and \(z_{j'}\in[0,u_{z,j'}]\) using the McCormick envelope:
\begin{equation}\label{eq:mccormick-zz}
\begin{aligned}
\sigma_{jj'} &\ge 0,\\
\sigma_{jj'} &\le u_{z,j'} z_j,\\
\sigma_{jj'} &\ge u_{z,j'}z_j + u_{z,j}z_{j'} - u_{z,j}u_{z,j'},\\
\sigma_{jj'} &\le u_{z,j} z_{j'},
\quad (j,j')\in\mathcal S .
\end{aligned}
\end{equation}
We also couple \(\sigma_{jj'}\) to the diagonal surrogates \(\zeta_j\) and
\(\zeta_{j'}\) through the corresponding local \(2\times2\) PSD minor, as
in~\eqref{eq:soc-minors}, equivalently imposing \(\zeta_j,\zeta_{j'}\ge0\) and
\(\sigma_{jj'}^2\le\zeta_j\zeta_{j'}\) for \((j,j')\in\mathcal S\).

Previous-layer couplings select products \(z_p^{(k-1)}z_q^{(k-1)}\), as in Figure~\ref{fig:SOCP}c. In the derivation here, \(k=1\) and \(z^{(0)}=x\). Let \(\mathcal T\subseteq[d]\times[d]\) with
\(p<q\), and introduce an auxiliary variable \(\tau_{pq}\) for each
\((p,q)\in\mathcal T\) to represent \(x_p x_q\). Given bounds
\(x_p\in[\ell_{x,p},u_{x,p}]\) and
\(x_q\in[\ell_{x,q},u_{x,q}]\), we impose the McCormick envelope
\citep{mccormick1976computability}:
\begin{equation}\label{eq:mccormick-xx}
\begin{aligned}
\tau_{pq} &\ge \ell_{x,p}x_q+\ell_{x,q}x_p-\ell_{x,p}\ell_{x,q},\\
\tau_{pq} &\ge u_{x,p}x_q+u_{x,q}x_p-u_{x,p}u_{x,q},\\
\tau_{pq} &\le u_{x,p}x_q+\ell_{x,q}x_p-u_{x,p}\ell_{x,q},\\
\tau_{pq} &\le \ell_{x,p}x_q+u_{x,q}x_p-\ell_{x,p}u_{x,q},
\quad (p,q)\in\mathcal T .
\end{aligned}
\end{equation}

We also couple \(\tau_{pq}\) to the diagonal surrogates \(\xi_p\) and
\(\xi_q\) through the corresponding local \(2\times2\) PSD minor, as
in~\eqref{eq:soc-minors}, equivalently imposing \(\xi_p,\xi_q\ge0\) and
\(\tau_{pq}^2\le\xi_p\xi_q\) for \((p,q)\in\mathcal T\).

\paragraph{Sparse Coupling Selection.}\leavevmode
At each hidden layer \(k\), the sparse sets \(\mathcal E^{(k)}\), \(\mathcal S^{(k)}\), and \(\mathcal T^{(k)}\) select cross-layer, current-layer, and previous-layer products, respectively. In our experiments, the selection rules are fixed across the evaluation, while the selected pairs are determined
separately for each input and target class. For each hidden layer \(k\), we
rank each unstable neuron \(j\) by
\begin{equation}\label{sparse-coupling-selection}
    \rho_j^{(k)}=
    |\gamma_j^{(k)}|\bigl(u_{s,j}^{(k)}-\ell_{s,j}^{(k)}\bigr), \quad \ell_{s,j}^{(k)}<0<u_{s,j}^{(k)},
\end{equation}
where \(\gamma_j^{(k)}\) is its target-dependent backward influence
coefficient. These scores rank current-layer candidates; interval widths rank previous-layer candidates. Pairs from these pools define the candidate couplings:
\(\mathcal E^{(k)}\), \(\mathcal S^{(k)}\), and
\(\mathcal T^{(k)}\). The backward recursion and the exact
pair-selection scores are given in
Appendix~\ref{app:sparse-coupling-selection}.

\begin{lemma}[Sparse SOCP Upper Bound]\label{lem:level1-upper} Let $\mathcal F_{\mathrm{SOCP}}(\bar x;\mathcal E,\mathcal S,\mathcal T)$ denote the feasible set of the sparse SOCP relaxation defined above, with optimal value
\begin{equation}\label{eq:ub-socp-def}
\begin{gathered}
f^{u}_{\mathrm{SOCP},y}(\bar{x})
\triangleq
\max \ c^\top z\\
\text{s.t.}\quad
(x,s,z,\xi,\zeta,\eta,\sigma,\tau,r)\in
\mathcal F_{\mathrm{SOCP}}(\bar{x};\mathcal E,\mathcal S,\mathcal T).
\end{gathered}
\end{equation}
Then
\begin{equation}\label{eq:socp-upper-bound}
f^\star_y(\bar{x})
\le
f^{u}_{\mathrm{SOCP},y}(\bar{x}).
\end{equation}
With unchanged variable bounds, enlarging $\mathcal E$, $\mathcal S$, or $\mathcal T$ cannot increase this upper bound. The full formulation and proof are given in Appendices~\ref{appendix:full-op} and~\ref{app:proof-socp-upper-bound}.
\end{lemma}

With the same variable bounds, SOCP adds selected coupling constraints to the LP relaxation. If SDP uses the same linear constraints as SOCP and imposes $P\succeq0$ on the full lifted matrix, the margin upper bounds satisfy \(f^{u}_{\mathrm{SDP},y}(\bar{x})\le
f^{u}_{\mathrm{SOCP},y}(\bar{x})\le
f^{u}_{\mathrm{LP},y}(\bar{x})\).
Consequently, on the same evaluation set, the corresponding certified robust accuracies satisfy
\(RA_{\mathrm{SDP}}\ge RA_{\mathrm{SOCP}}\ge RA_{\mathrm{LP}}\).

\subsection{Extension to Multiple Hidden Layers}\label{rem:multilayer}
For \(K\) hidden layers, let \(z^{(0)}=x\), \(s^{(k)}=W^{(k)}z^{(k-1)}\), and \(z^{(k)}=\operatorname{ReLU}(s^{(k)})\).
At layer \(k\), \(z^{(k-1)}\) takes the role of \(x\). The sets \(\mathcal E^{(k)}\), \(\mathcal S^{(k)}\), and \(\mathcal T^{(k)}\) select products \(z_i^{(k-1)}z_j^{(k)}\), \(z_j^{(k)}z_{j'}^{(k)}\), and \(z_p^{(k-1)}z_q^{(k-1)}\), respectively.
Thus \(\mathcal T^{(1)}\) represents original input-input products, whereas \(\mathcal T^{(k)}\) for \(k\ge2\) represents products of preceding-layer hidden activations. These constraints are valid in both cases, even when they leave the margin bound unchanged.

We apply the interval bounds, ReLU hulls, McCormick envelopes, and local PSD constraints at each layer, sharing square lifts across incident products and adjacent layers, with \(\zeta^{(0)}=\xi\). Appendix~\ref{sec:socp-multilayer} gives the multilayer residual identity, output margin, residual bounds, and sparse implementation.

\begin{remark}[Computational Complexity]\label{rem:complexity}
Let \(Q=|\mathcal E|+|\mathcal S|+|\mathcal T|\).
The SOCP has \(O(d+m+Q)\) scalar variables and constraints, compared
with \(O((d+m)^2)\) entries in the dense SDP lift.
A conventional interior-point solve costs
\(O(I_{\mathrm{SOCP}}M^3)\) arithmetic operations with dense
factorization, where \(M=O(d+m+Q)\) is the Newton-system dimension
and \(I_{\mathrm{SOCP}}\) the iteration count. Appendix~\ref{appendix:time_complexity} details construction costs,
memory, multilayer scaling, and the comparison with branch-and-bound.
\end{remark}

\section{CORE: Class-wise Cascade}\label{sec:core}
For an input $\bar{x}$ with a true label $\bar{y}$, certification reduces to $L\!-\!1$ class-wise margin problems: $\bar{x}$ is certified robust if, for every $y\neq\bar{y}$, an upper bound \(f^{u}_{y}(\bar{x})\) is negative. 
CRV escalates inputs across verifiers and may re-solve all class-wise subproblems at expensive stages~\citep{maleki2026cascading}. 
CORE cascades at the \textit{class-subproblem} level: for each target class, it applies verifiers from cheap to expensive and escalates only when the current stage returns a non-negative upper bound for that class (Algorithm~\ref{alg:core}).

If the upper bound for a class remains nonnegative at the final stage, CORE terminates immediately for that input (Algorithm~\ref{alg:core}), because the cascade cannot certify the input as robust.

\begin{algorithm}[t]
\caption{CORE: Class-wise Verification}\label{alg:core}
\begingroup
\algrenewcommand{\algorithmiccomment}[1]{\hfill\(\triangleright\)\,#1}
\begin{algorithmic}[1]
\Require Clean input $\bar{x}$, true label $\bar{y}$, radius $\varepsilon$
\Require Ordered verifiers $\mathcal{V}_1,\dots,\mathcal{V}_H$
\Ensure \textsc{Robust} or \textsc{Uncertified}
\For{each $y\in[L]$ with $y\neq \bar{y}$}\label{ln:core-classloop}
    \For{$h=1$ to $H$}\label{ln:core-stageloop}
        \State $U \gets \mathcal{V}_h(\bar{x},\bar{y},y,\varepsilon)$ \Comment{upper bound on margin}\label{ln:core-eval}
        \If{$U < 0$}
            \State \textbf{break}\Comment{class $y$ certified}\label{ln:core-cert}
        \EndIf
        \If{$h=H$}
            \State \Return \textsc{Uncertified}\label{ln:core-fail}
        \EndIf
    \EndFor\label{ln:core-escalate}
\EndFor
\State \Return \textsc{Robust}\label{ln:core-success}
\end{algorithmic}
\endgroup
\end{algorithm}

We assume a fixed ordering of verifiers $\mathcal{V}_1,\dots,\mathcal{V}_H$ in increasing order of typical computational cost, based on their relaxation family. In particular, linear relaxations (e.g., bound propagation/LP-style) are placed earlier, SOCP relaxations follow, and semidefinite (or other heavier quadratic/SDP-based) relaxations are placed last. This ordering is chosen \emph{once} and used for all inputs and classes, based on theoretical and empirical complexity rather than input-dependent heuristics (see, e.g., the solver-level cost summaries in~\citep{li2023sok}). When the relative cost of two candidates is unclear, we fix their order using a brief pilot run on a small representative subset; if their costs are comparable, we prioritize the tighter verifier first, since early certification avoids invoking later stages.

CORE certifies an input as robust only when every adversarial class is certified with a negative upper bound by at least one verifier; otherwise, it stops as soon as a class remains non-negative even under the tightest verifier $\mathcal{V}_H$. Consequently, CORE’s certified robust accuracy is at least that of  $\mathcal{V}_H$, as follows from the cascade logic in Algorithm~\ref{alg:core}. 

CORE can reduce runtime by avoiding expensive evaluations for classes certified at earlier stages and by stopping immediately once the cascade cannot certify an input.

%% file: sections/Experiment.tex
\section{Experiments}\label{sec:experiment}
Experiments on MNIST and CIFAR-10 under \(\ell_\infty\) perturbations address three questions: (i) How do SOCP-, LP-, and SDP-based verifiers compare in certified robust accuracy and runtime? (ii) Can CORE reduce runtime while retaining the final verifier's certified robust accuracy? (iii) How does the SOCP-based verifier compare with \(\alpha,\beta\)-CROWN across certified-training methods? 

Computing exact robust accuracy is generally intractable. Certified robust accuracy and accuracy under PGD attacks provide lower and upper bounds, respectively. We use \(\varepsilon_{\mathrm{train}}\) for the training perturbation budget and \(\varepsilon_{\mathrm{verify}}\) for the verification budget, with \(\varepsilon\) as shorthand for the latter.

\begin{table*}[t]
  \centering
  \normalsize
  \setlength{\abovecaptionskip}{\baselineskip}
  \setlength{\belowcaptionskip}{\baselineskip}
  \setlength{\tabcolsep}{3pt}
  \renewcommand{\arraystretch}{1.05}
  \caption{Verification of CNNs Trained for Certified Robustness. Clean accuracy, certified robust accuracy (SOCP-cert and \(\alpha,\beta\)-CROWN), and PGD upper bounds use matched test images (300 MNIST; 200 CIFAR-10). N/A indicates that \(\alpha,\beta\)-CROWN ran out of memory (Appendix~\ref{app:resource-limited-verification}). Additional MNIST results at \(\varepsilon_{\mathrm{verify}}=0.2\) appear in Appendix~\ref{app:additional-results}.}
  \label{tab:robust-trained-ab-crown}
  \begin{tabular*}{\textwidth}{@{\extracolsep{\fill}}llrrrrrr@{}}
    \toprule
    \textbf{Dataset} & \textbf{Training Method} &
    \textbf{\boldmath\(\varepsilon_{\mathrm{train}}\)} &
    \textbf{\boldmath\(\varepsilon_{\mathrm{verify}}\)} &
    \textbf{Clean} & \textbf{SOCP-cert} &
    \textbf{\boldmath\(\alpha,\beta\)-CROWN} & \textbf{PGD} \\
    \midrule
    \multirow{4}{*}{MNIST}
      & IBP       & 0.1 & 0.1 & 99.00\% & 88.33\% & 87.33\% & 88.67\% \\
      & CROWN-IBP & 0.1 & 0.1 & 97.67\% & 94.33\% & 93.67\% & 94.67\% \\
      & DeepPoly  & 0.1 & 0.1 & 98.00\% & 94.67\% & 93.33\% & 94.67\% \\
      & Dual-WK   & 0.1 & 0.1 & 82.00\% & 69.67\% & 70.00\% & 72.33\% \\
    \midrule
    \multirow{2}{*}{MNIST}
      & CROWN-IBP & 0.1 & 0.15 & 97.67\% & 8.00\% & N/A & 32.00\% \\
      & DeepPoly  & 0.1 & 0.15 & 98.00\% & 7.33\% & N/A & 34.33\% \\
    \midrule
    \multirow{4}{*}{MNIST}
      & IBP       & 0.3 & 0.3 & 99.00\% & 71.67\% & 70.33\% & 78.33\% \\
      & CROWN-IBP & 0.3 & 0.3 & 97.67\% & 83.33\% & 83.33\% & 85.67\% \\
      & DeepPoly  & 0.3 & 0.3 & 98.00\% & 82.33\% & 82.00\% & 85.67\% \\
      & Dual-WK   & 0.3 & 0.3 & 82.00\% & 51.00\% & 48.33\% & 51.33\% \\
    \midrule
    \multirow{3}{*}{CIFAR-10}
      & CROWN-IBP & 2/255 & 1/255 & 54.00\% & 50.00\% & 49.00\% & 51.00\% \\
      & DeepPoly  & 2/255 & 1/255 & 59.50\% & 49.00\% & 49.00\% & 51.00\% \\
      & Dual-WK   & 2/255 & 1/255 & 42.00\% & 38.50\% & 38.00\% & 39.00\% \\
    \midrule
    \multirow{3}{*}{CIFAR-10}
      & CROWN-IBP & 2/255 & 2/255 & 54.00\% & 39.50\% & 39.50\% & 39.50\% \\
      & DeepPoly  & 2/255 & 2/255 & 59.50\% & 36.50\% & 35.50\% & 38.00\% \\
      & Dual-WK   & 2/255 & 2/255 & 42.00\% & 29.00\% & 28.50\% & 31.50\% \\
    \bottomrule
  \end{tabular*}
  \par\vspace{\baselineskip}
\end{table*}

\noindent\textbf{Networks.}
Table~\ref{tab:robust-trained-ab-crown} compares SOCP-cert with \(\alpha,\beta\)-CROWN across IBP~\citep{gowal2018ibp}, CROWN-IBP~\citep{zhang2020crownibp}, DeepPoly~\citep{singh2019deeppoly}, and Dual-WK~\citep{wong2018provable} training, using a fixed architecture per dataset: a LeNet-style CNN for MNIST and CNN5 for CIFAR-10.

We compare certified robust accuracy and runtime on matched networks and test inputs using \textit{Grad-NN}, trained with SDP-based robust regularization, and \textit{Std-NN}, trained by empirical risk minimization. \textit{CNNSmall-A} (MNIST) and \textit{CNNSmall-B} (CIFAR-10) extend the LP/SOCP comparison to CNNs; Grad-NN also evaluates CORE's runtime savings. Architecture and training details appear in Appendix~\ref{app:experimental-networks}.

\textbf{Certification Procedures.} LP-cert uses the linear relaxation of~\citet{wong2018provable}; SOCP-cert uses the sparse lifted relaxation developed in Section~\ref{sec:pure-socp}; and SDP-cert uses a global PSD lift~\citep{raghunathan2018semidefinite}. \(\alpha,\beta\)-CROWN combines bound propagation with branch-and-bound~\citep{wang2021beta}.

CRV-FSR1 (\(\mathrm{LP}\to\mathrm{SDP}\)) and CRV-FSR2 (\(\mathrm{SOCP}\to\mathrm{SDP}\)) cascade at the input level with fast stepwise relaxation (FSR)~\citep{maleki2026cascading}. CORE applies \(\mathrm{LP}\to\mathrm{SOCP}\to\mathrm{SDP}\) at the target-class level (Algorithm~\ref{alg:core}); CORE-FSR adds FSR to its SDP stage.

\begin{figure}[t]
    \centering
    \includegraphics[width=0.70\columnwidth]{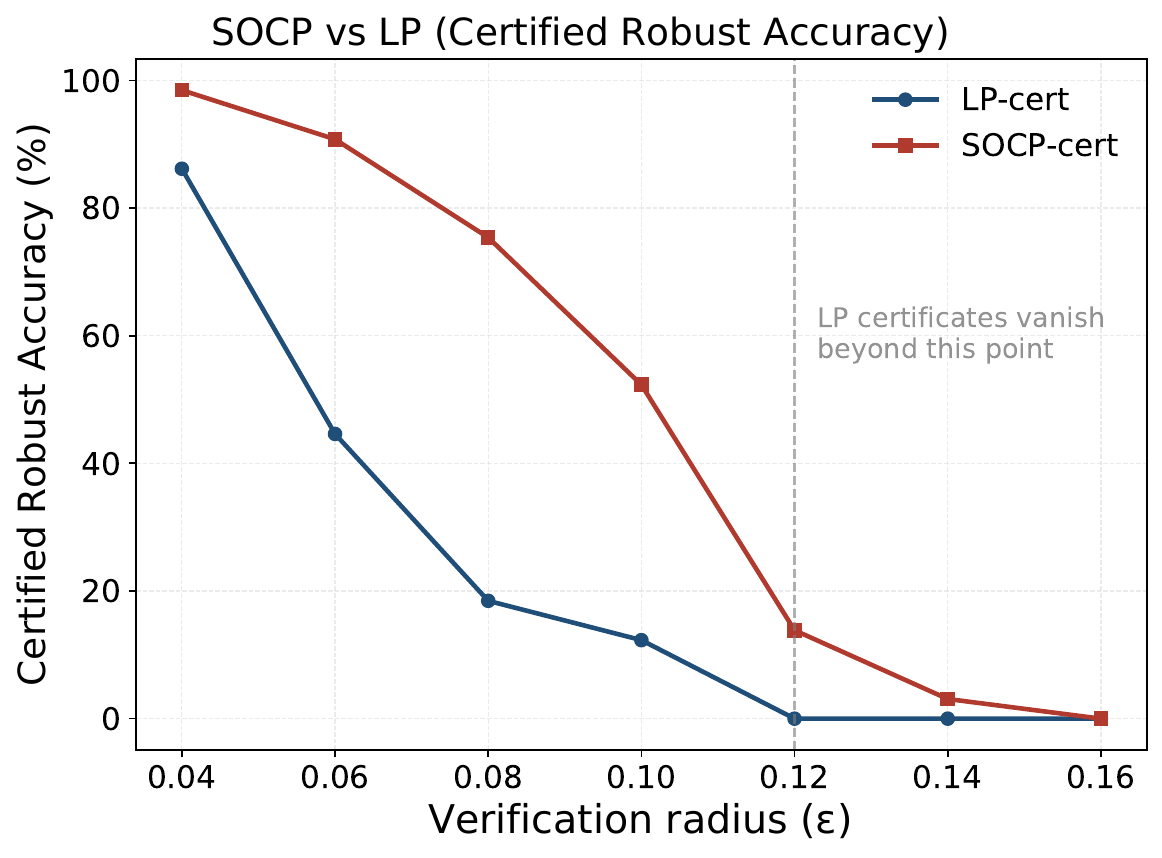}
    \caption{Certified robust accuracy versus \(\ell_\infty\) perturbation radius on Grad-NN (one hidden layer). SOCP-cert retains certificates at radii where LP-cert certifies no inputs.}
    \label{fig:epsilon-1layer}
    \label{fig:epsilon-combined}
\end{figure}

\begin{figure}[t]
    \centering
    \includegraphics[width=0.70\columnwidth]{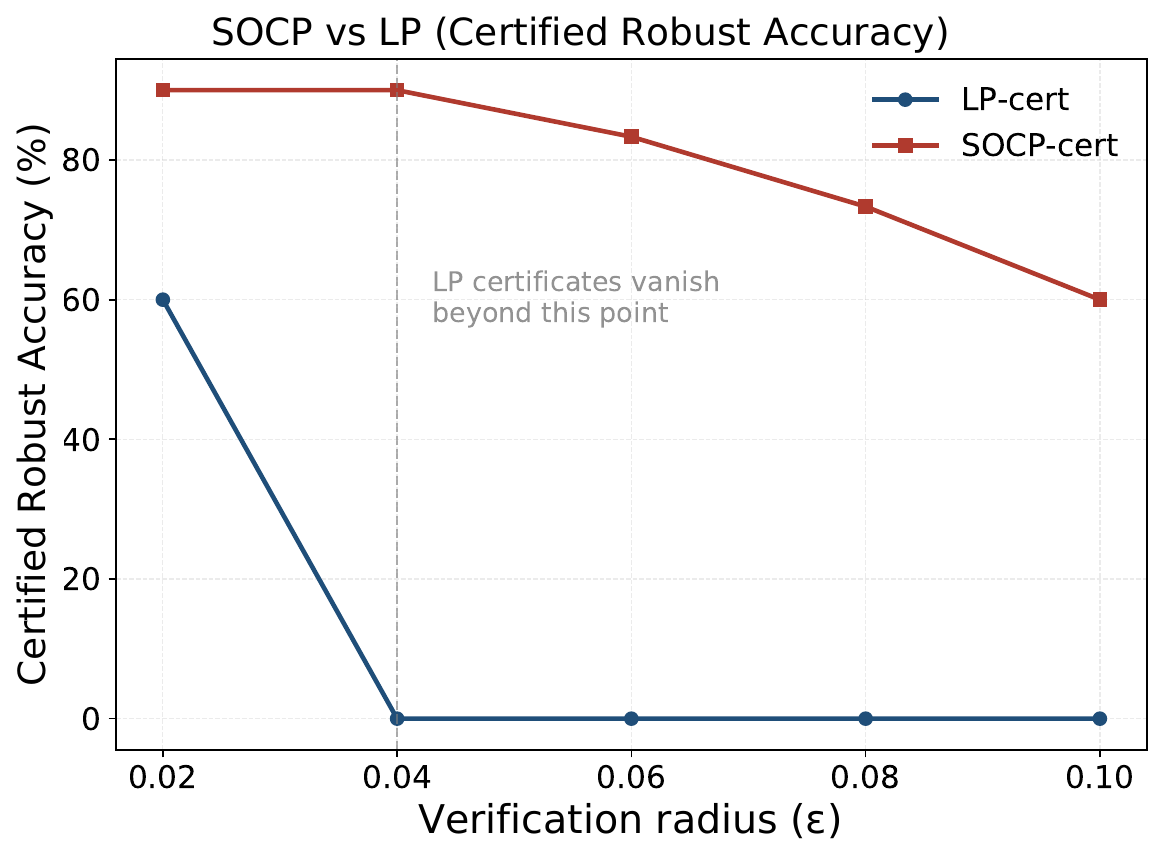}
    \caption{Certified robust accuracy versus \(\ell_\infty\) perturbation radius on Std-NN (two hidden layers). SOCP-cert improves on LP-cert across the evaluated radii.}
    \label{fig:epsilon-2layer}
\end{figure}

\noindent\textbf{Optimization setup.}
For Table~\ref{tab:robust-trained-ab-crown}, SOCP-cert uses \texttt{CVXPY} (SCS or CLARABEL); \(\alpha,\beta\)-CROWN uses CUDA with a 120-second per-property timeout and a 10,000-domain limit. For Table~\ref{tab:CORE} and Figures~\ref{fig:epsilon-1layer}--\ref{fig:epsilon-2layer}, SOCP-cert and SDP-cert use \texttt{YALMIP}~\citep{lofberg2004yalmip}/\texttt{MOSEK} on the same 6-core CPU, and LP-cert uses \texttt{PyTorch}\footnote{The source code is available at \url{https://github.com/SadigDaniel/SOCP-Verifier}.}. Appendix~\ref{app:experimental-optimization} gives additional solver settings.

\paragraph{SOCP-cert selected couplings.}
For the layer-wise coupling sets defined in Section~\ref{rem:multilayer}, the baseline budgets \((B_E,B_S,B_T)=(16,8,8)\) retain at most 32 products per hidden layer. Current-layer candidates are ranked by the score defined in~\eqref{sparse-coupling-selection}, which is the product of each unstable ReLU's preactivation interval width and its target-dependent influence. We retain the highest-scoring eligible pairs separately for each input and target class. All three sets are imposed jointly through McCormick envelopes and SOC-representable \(2\times2\) PSD constraints. Exact pair scores are given in Appendix~\ref{app:sparse-coupling-selection}.

\paragraph{SOCP-cert vs. \(\alpha,\beta\)-CROWN.}
In Table~\ref{tab:robust-trained-ab-crown}, SOCP-cert achieves higher mean certified robust accuracy than \(\alpha,\beta\)-CROWN on both datasets, with comparable average runtime on MNIST (43.3 versus 40 seconds per image); on CIFAR-10, its runtime is higher (approximately 305 versus 120 seconds).

Certified training can reduce ReLU instability~\citep{mao2024ctbench}, favoring branch-and-bound. Increasing the verification radius, however, can introduce additional unstable ReLUs and make branching more demanding. For the CROWN-IBP and DeepPoly MNIST models trained at \(\varepsilon_{\mathrm{train}}=0.1\) and verified at \(\varepsilon_{\mathrm{verify}}=0.15\), \(\alpha,\beta\)-CROWN runs out of available memory, whereas SOCP-cert completes both 300-image evaluations, certifying 8.00\% and 7.33\%, respectively. This highlights a practical advantage of SOCP-cert's fixed relaxation: by avoiding a growing search frontier, it completes under the tested memory budget (Appendix~\ref{app:resource-limited-verification}).

\begin{table}[t]
    \centering
    \normalsize
    \setlength{\abovecaptionskip}{\baselineskip}
    \setlength{\belowcaptionskip}{\baselineskip}
    \setlength{\tabcolsep}{3pt}
    \renewcommand{\arraystretch}{1.05}
    \caption{Verification of Grad-NN on MNIST at \(\varepsilon=0.1\). RA is certified robust accuracy; PGD gives an empirical upper bound. Time is averaged per image; reduction is relative to SDP-cert.}
    \label{tab:CORE}
    \begin{tabular*}{\columnwidth}{@{\extracolsep{\fill}}lrrr@{}}
      \toprule
      \textbf{Verifier} & \textbf{RA} & \textbf{Time (min)} & \textbf{Reduction}\\
      \midrule
      SDP-cert & 86.15\% & 328.98 & --\\
      SOCP-cert & 52.31\% & 0.31 & 99.90\%\\
      LP-cert & 12.31\% & -- & --\\
      \midrule
      CRV-FSR1 & 86.15\% & 168.33 & 48.83\%\\
      CRV-FSR2 & 86.15\% & 91.86 & 72.08\%\\
      CORE & 86.15\% & 70.61 & 78.54\%\\
      CORE-FSR & 86.15\% & 45.06 & 86.30\%\\
      \midrule
      PGD & 90.00\% & -- & --\\
      \bottomrule
    \end{tabular*}
    \par\vspace{\baselineskip}
\end{table}

\paragraph{SOCP-cert vs. SDP-cert.}
SOCP-cert offers an intermediate tightness-cost trade-off: selected second-order couplings strengthen LP-cert, while the sparse lift uses small SOC constraints instead of SDP-cert's global PSD constraint, reducing formulation size at the evaluated coupling budgets (Appendix~\ref{appendix:time_complexity}). On Grad-NN at \(\varepsilon=0.1\) (Table~\ref{tab:CORE}), SOCP-cert certifies 52.31\% of inputs, compared with 12.31\% for LP-cert and 86.15\% for SDP-cert. Its mean runtime is 0.31 versus 328.98 minutes per image for SDP-cert, demonstrating substantially lower computational cost at the expense of lower certified robust accuracy on this benchmark.

\paragraph{SOCP-cert vs. LP-cert.} We also apply both verifiers to CNNs on both datasets: compared to LP-cert, SOCP-cert improves certified robust accuracy from 0.00\% to 86.67\% on MNIST CNNSmall-A at \(\varepsilon=0.010\), and from 2.00\% to 41.02\% on CIFAR-10 CNNSmall-B at \(\varepsilon=0.001\).

\paragraph{CORE.}
All four cascades in Table~\ref{tab:CORE} attain SDP-cert's 86.15\% certified robust accuracy. Using SOCP-cert rather than LP-cert as the first stage reduces CRV-FSR runtime from 168.33 to 91.86 minutes per image. CORE takes 70.61 minutes, and CORE-FSR takes 45.06 minutes, an 86.30\% reduction relative to SDP-cert. CORE-FSR also improves on CRV-FSR2 (45.06 versus 91.86 minutes) at the same certified robust accuracy. PGD yields a 90.00\% robust-accuracy upper bound; the remaining gap to CORE is 3.85 percentage points.

\paragraph{Effect of perturbation level $\varepsilon$.}
Figures~\ref{fig:epsilon-1layer} and~\ref{fig:epsilon-2layer} compare SOCP-cert and LP-cert on the fully connected networks. On Grad-NN, with one hidden layer, SOCP-cert certifies 13.85\% at \(\varepsilon=0.12\), whereas LP-cert reaches 0.00\%. On Std-NN, with two hidden layers, LP-cert reaches 0.00\% at 0.04, whereas SOCP-cert still certifies 60.00\% at 0.10.

%% file: sections/Conclusion.tex
\section{Discussion and Limitations}\label{sec:conclusion}

We presented an SOCP-based certification framework for ReLU networks that bridges the gap between LP- and SDP-based verification. Starting from a QCQP formulation of the robustness problem, our relaxation captures selected second-order structure through sparse lifting, McCormick envelopes, and SOC-representable \(2\times2\) PSD minors. We further introduced CORE, a class-subproblem cascade that escalates only unresolved margins to tighter verifiers, reducing redundant computation while preserving the certified robust accuracy of the tightest verifier in the cascade. Limitations include that the certificates provide deterministic guarantees for the specified \(\ell_\infty\) threat model, and that the tightness-runtime trade-off depends on the selected sparse coupling sets. 

%% file: sections/Appendix.tex
\section{Experimental Setup and Additional Results}
\label{app:experimental-details}

\subsection{Networks}
\label{app:experimental-networks}
\begin{itemize}
\item \textit{Grad-NN}: a fully connected network with one 500-unit hidden layer, trained using SDP-based robust regularization~\citep{raghunathan2018semidefinite}.
\item \textit{Std-NN}: a fully connected network with two hidden layers, trained by standard empirical risk minimization.
\item \textit{CNNSmall-A}: a MNIST network with \(1\times28\times28\) inputs, two \(3\times3\) convolution-ReLU layers with 8/16 channels, a 64-unit fully connected ReLU layer, and a linear classifier; approximately 0.8 million parameters.
\item \textit{CNNSmall-B}: the same layer structure as CNNSmall-A, with \(3\times32\times32\) CIFAR-10 inputs; approximately 1.05 million parameters.
\item \textit{LeNet-style CNN}: the certified-trained MNIST architecture in Table~\ref{tab:robust-trained-ab-crown}, with two convolution-ReLU blocks (16/32 channels, \(4\times4\) kernels, stride 2, padding 1), a 100-unit fully connected ReLU layer, and a 10-class linear output; 166,406 trainable parameters.
\item \textit{CNN5}: the certified-trained CIFAR-10 architecture in Table~\ref{tab:robust-trained-ab-crown}, with five convolution-ReLU blocks (24/24/48/48/96 channels), a 192-unit fully connected ReLU layer, and a 10-class linear output; 416,506 trainable parameters. Blocks 1/3/5 use \(4\times4\) kernels and stride 2; blocks 2/4 use \(3\times3\) kernels and stride 1; all use padding 1.
\end{itemize}

\paragraph{Training data and preprocessing.}
Inputs use pixels in \([0,1]\) without mean/std normalization. MNIST uses no augmentation; CIFAR-10 training uses random horizontal flips and crops with four-pixel padding. Hyperparameters are selected on a held-out 10\% training split. Verification uses unaugmented inputs in evaluation mode, with perturbations clipped to the pixel range.

\paragraph{Training settings.}
The certified-trained models use AdamW with batch size 128 and an initial learning rate of \(10^{-3}\). MNIST models are trained for 20 epochs with weight decay \(10^{-4}\); CIFAR-10 models for 35 epochs with weight decay \(5\times10^{-4}\).

\subsection{Certification and Optimization Details}
\label{app:experimental-optimization}

\paragraph{Coupling budgets.}
For Table~\ref{tab:robust-trained-ab-crown}, the per-layer budgets are \((B_E,B_S,B_T)=(16,8,8)\), as specified in Section~\ref{sec:experiment}. Appendix~\ref{app:sparse-coupling-selection} gives the candidate selection and pair scores. Candidate-pool limits control the pairs considered for scoring; the coupling budgets limit the pairs retained in the conic problem.

\paragraph{Solver choice.}
The software and hardware for the two experimental setups are specified in Section~\ref{sec:experiment}. For challenging CVXPY instances, CLARABEL is used to improve numerical convergence and accuracy; changing between SCS and CLARABEL preserves the relaxation's mathematical feasible set. Resource-limited runs are discussed in Appendix~\ref{app:resource-limited-verification}.

\paragraph{\(\alpha,\beta\)-CROWN configuration.}
The runs use an NVIDIA TITAN Xp with 12\,GiB of GPU memory and approximately 15\,GiB of system RAM. Incomplete verification precedes branch-and-bound, with patch-mode convolution and KFSB branching. The timeout and domain limit are specified in Section~\ref{sec:experiment}. The initial batch size is 1024 with automatic enlargement disabled; after premature termination, evaluation is repeated with smaller batch sizes. We use 100 \(\alpha\)-CROWN optimization iterations and 20 \(\beta\)-CROWN iterations. The internal PGD attack, when enabled, runs before certification and is distinct from the separate PGD evaluation reported in the tables.

\subsection{Additional Results}
\label{app:additional-results}

\begin{table}[t]
  \centering
  \caption{Certified-trained MNIST models trained at \(\varepsilon_{\mathrm{train}}=0.3\) and evaluated at \(\varepsilon_{\mathrm{verify}}=0.2\), complementing Table~\ref{tab:robust-trained-ab-crown}. Accuracy values are percentages on the same matched 300-image evaluation set.}
  \label{tab:robust-trained-mnist-eps02}
  \resizebox{\linewidth}{!}{%
  {
  \begin{tabular}{llrrrrrr}
    \toprule
    Dataset & Training method & \(\varepsilon_{\mathrm{train}}\) &
    \(\varepsilon_{\mathrm{verify}}\) & Clean & SOCP-cert &
    \(\alpha,\beta\)-CROWN & PGD \\
    \midrule
    \multirow{4}{*}{MNIST}
      & IBP       & 0.3 & 0.2 & 99.00 & 83.33 & 81.67 & 86.00 \\
      & CROWN-IBP & 0.3 & 0.2 & 97.67 & 88.33 & 87.33 & 90.00 \\
      & DeepPoly  & 0.3 & 0.2 & 98.00 & 88.00 & 86.67 & 90.33 \\
      & Dual-WK   & 0.3 & 0.2 & 82.00 & 61.67 & 61.00 & 62.67 \\
    \bottomrule
  \end{tabular}%
  }}
\end{table}

\paragraph{Certified-trained models at verification radius 0.2.}
Table~\ref{tab:robust-trained-mnist-eps02} reports the MNIST results at verification radius 0.2. SOCP-cert exceeds \(\alpha,\beta\)-CROWN on all four reported models by 0.67--1.67 percentage points, while remaining below the PGD robust-accuracy upper bounds. These observations concern the evaluated models, images, and configurations.

\section{Residual Bounds}
\label{app:residual-bounds}

Write \(W=W_1\) for the one-hidden-layer network. For each hidden unit \(j\), the residual in~\eqref{eq:zeta-residual-main} collects the omitted products:
\[
r_j=\sum_{i:(i,j)\notin\mathcal E}W_{ji}x_i z_j.
\]
For \(x_i\in[\ell_{x,i},u_{x,i}]\) and \(z_j\in[0,u_{z,j}]\), define
\[
\underline p_{ij}=u_{z,j}\min\{0,\ell_{x,i}\},
\qquad
\overline p_{ij}=u_{z,j}\max\{0,u_{x,i}\}.
\]
Then \(x_i z_j\in[\underline p_{ij},\overline p_{ij}]\), and signed interval arithmetic gives
\begin{equation}\label{eq:residual-interval-appendix}
\begin{aligned}
L_j^{(r)}&=\sum_{i:(i,j)\notin\mathcal E}
\min\{W_{ji}\underline p_{ij},W_{ji}\overline p_{ij}\},\\
U_j^{(r)}&=\sum_{i:(i,j)\notin\mathcal E}
\max\{W_{ji}\underline p_{ij},W_{ji}\overline p_{ij}\}.
\end{aligned}
\end{equation}
Thus \(L_j^{(r)}\le r_j\le U_j^{(r)}\). Empty sums are zero, so retaining every cross-layer product eliminates the residual. The bounds are conservative because they need not capture dependencies among omitted products.

At hidden layer \(k\), set \(a=z_i^{(k-1)}\), \(b=z_j^{(k)}\), \(W=W^{(k)}\), and \(\mathcal E=\mathcal E^{(k)}\). For bounds \(a\in[\ell_a,u_a]\) and \(b\in[\ell_b,u_b]\), use
\[
\underline p=\min\{\ell_a\ell_b,\ell_a u_b,u_a\ell_b,u_a u_b\},
\qquad
\overline p=\max\{\ell_a\ell_b,\ell_a u_b,u_a\ell_b,u_a u_b\}.
\]
Substituting these endpoints into~\eqref{eq:residual-interval-appendix} gives valid layer-wise residual bounds, including when an activation has a positive lower bound.
\section{Full SOCP Formulation}\label{appendix:full-op}

For one hidden layer, the feasible set $\mathcal F_{\mathrm{SOCP}}(\bar x;\mathcal E,\mathcal S,\mathcal T)$ consists of the ReLU convex-hull constraints, diagonal lifts, McCormick envelopes for selected cross-layer, current-layer, and previous-layer products, residual bounds, and SOC-representable PSD constraints on selected $2\times2$ principal submatrices. Here \(z^{(0)}=x\), and layer superscripts are omitted; Appendix~\ref{sec:socp-multilayer} gives the multilayer formulation.
For fixed $(\bar{x},\bar{y},y)$, let $s=W_1x$, $W=W_1$, and \(c=W_2^\top(e_y-e_{\bar y})\). With the residual intervals from Appendix~\ref{app:residual-bounds}, the upper bound is given by
\begin{align}\label{eq:ub3-full}
\begin{split}
f^{u}_{\mathrm{SOCP},y}(\bar{x}) \triangleq\; &\max_{x,z,\xi,\zeta,\eta,\sigma,\tau,r}\quad c^\top z\\
\text{s.t.}\quad
& x \in [\ell_x,u_x],\qquad s = W_1x,\\[2pt]
& \forall j\in[m], \qquad z_j \ge 0,\qquad z_j \ge s_j,\\
&\qquad z_j \le \alpha_j(s_j-\ell_{s,j})\ \ \text{if }\ell_{s,j}<0<u_{s,j},\quad
\alpha_j=\frac{u_{s,j}}{u_{s,j}-\ell_{s,j}},\\
&\qquad z_j=s_j\ \ \text{if }\ell_{s,j}\ge 0,\qquad z_j=0\ \ \text{if }u_{s,j}\le 0,\\[2pt]
& \forall i\in[d],\qquad 0 \le \xi_i \le (\ell_{x,i}+u_{x,i})x_i - \ell_{x,i}u_{x,i},\\
&\qquad \zeta_j \ge 0,\ \forall j\in[m],\\[2pt]
& \forall (i,j)\in\mathcal{E}, \qquad \eta_{ij} \ge \ell_{x,i}\, z_j,\qquad \eta_{ij} \le u_{x,i}\, z_j,\\
&\qquad \eta_{ij} \ge u_{x,i}\, z_j + u_{z,j}\, x_i - u_{x,i}u_{z,j},\\
&\qquad \eta_{ij} \le \ell_{x,i}\, z_j + u_{z,j}\, x_i - \ell_{x,i}u_{z,j},\\[2pt]
& \forall (i,j)\in\mathcal{E}, \qquad x_i^2 \le \xi_i,\qquad z_j^2 \le \zeta_j,\qquad \eta_{ij}^2 \le \xi_i\zeta_j,\\[2pt]
& \forall j\in[m], \qquad \zeta_j = \sum_{i:(i,j)\in\mathcal{E}} W_{ji}\eta_{ij} + r_j,\\
&\qquad r_j \in [L^{(r)}_j,U^{(r)}_j],\\
& \forall (j,j')\in\mathcal{S},\ j<j', \qquad \sigma_{jj'} \ge 0,\qquad \sigma_{jj'} \le u_{z,j'} z_j,\qquad \sigma_{jj'} \le u_{z,j} z_{j'},\\
&\qquad \sigma_{jj'} \ge u_{z,j'} z_j + u_{z,j} z_{j'} - u_{z,j}u_{z,j'},\\
& \forall (j,j')\in\mathcal{S}, \qquad \sigma_{jj'}^2 \le \zeta_j\zeta_{j'},\\[2pt]
& \forall (p,q)\in\mathcal{T},\ p<q, \qquad \tau_{pq} \ge \ell_{x,p}x_q + \ell_{x,q}x_p - \ell_{x,p}\ell_{x,q},\\
&\qquad \tau_{pq} \ge u_{x,p}x_q + u_{x,q}x_p - u_{x,p}u_{x,q},\\
&\qquad \tau_{pq} \le u_{x,p}x_q + \ell_{x,q}x_p - u_{x,p}\ell_{x,q},\\
&\qquad \tau_{pq} \le \ell_{x,p}x_q + u_{x,q}x_p - \ell_{x,p}u_{x,q},\\
& \forall (p,q)\in\mathcal{T}, \qquad \tau_{pq}^2 \le \xi_p\xi_q.
\end{split}
\end{align}
\section{Proof of Lemma~\ref{lem:level1-upper}}
\label{app:proof-socp-upper-bound}

\paragraph{Validity.}
Take any exact feasible point \(x\in[\ell_x,u_x]\), \(s=W_1x\), and \(z=\operatorname{ReLU}(s)\). Complete it with
\[
\xi_i=x_i^2,\qquad \zeta_j=z_j^2,\qquad
\eta_{ij}=x_i z_j,\qquad \sigma_{jj'}=z_jz_{j'},\qquad
\tau_{pq}=x_px_q
\]
for the selected pairs, and set \(r_j=\sum_{i:(i,j)\notin\mathcal E}W_{ji}x_i z_j\).
The exact ReLU graph satisfies its convex hull. The input square secant follows from
\((x_i-\ell_{x,i})(x_i-u_{x,i})\le0\), and every exact product satisfies its McCormick envelope. For any selected pair \((a,b)\), its lifted minor is
\[
\begin{bmatrix}a^2&ab\\ab&b^2\end{bmatrix}
=\begin{bmatrix}a\\b\end{bmatrix}
\begin{bmatrix}a\\b\end{bmatrix}^{\!\top}\succeq0,
\]
so all SOC constraints hold. Finally, ReLU complementarity gives
\[
z_j^2=z_js_j=\sum_{i:(i,j)\in\mathcal E}W_{ji}\eta_{ij}+r_j,
\]
and Appendix~\ref{app:residual-bounds} bounds the omitted sum. Thus every exact feasible point has a feasible lifted completion in~\eqref{eq:ub3-full}. Projecting onto \((x,z)\) and maximizing the same objective proves
\(f_y^\star(\bar x)\le f^u_{\mathrm{SOCP},y}(\bar x)\).

\paragraph{Monotonicity.}
Fix the variable bounds and take
\(\mathcal E_1\subseteq\mathcal E_2\),
\(\mathcal S_1\subseteq\mathcal S_2\), and
\(\mathcal T_1\subseteq\mathcal T_2\).
Given a feasible point for the larger sets, retain the shared variables and define
\[
r_j^{(1)}=r_j^{(2)}+
\sum_{i:(i,j)\in\mathcal E_2\setminus\mathcal E_1}W_{ji}\eta_{ij}.
\]
The McCormick envelope bounds each newly selected product within the interval previously used for that omitted term. Summing those intervals with the remaining residual interval places \(r_j^{(1)}\) in the bounds for \(\mathcal E_1\), and the residual equation is preserved. All other constraints on shared variables remain in the larger formulation. Therefore
\[
\operatorname{proj}_{x,z}\mathcal F_{\mathrm{SOCP}}(\bar x;\mathcal E_2,\mathcal S_2,\mathcal T_2)
\subseteq
\operatorname{proj}_{x,z}\mathcal F_{\mathrm{SOCP}}(\bar x;\mathcal E_1,\mathcal S_1,\mathcal T_1),
\]
which proves that enlarging the selected sets cannot increase the maximum. This is a non-strict statement: additional couplings need not improve the bound. Replacing existing pairs or changing the underlying variable bounds is not covered by this argument.

\paragraph{Comparison with LP and SDP.}
SOCP retains the input bounds and ReLU hulls of the corresponding LP, hence its projection is contained in the LP feasible set. Conversely, a full lifted matrix satisfying the same linear constraints and global positive semidefiniteness satisfies every selected PSD minor and square epigraph. For these matched formulations,
\[
f_y^\star(\bar x)\le f^u_{\mathrm{SDP},y}(\bar x)
\le f^u_{\mathrm{SOCP},y}(\bar x)
\le f^u_{\mathrm{LP},y}(\bar x).
\]
This comparison requires the same bounds and retained linear constraints; it does not order arbitrary implementations with different preprocessing or numerical settings.

\section{Extension to Multiple Hidden Layers}
\label{sec:socp-multilayer}

\paragraph{Network and interval bounds.}
For \(K\) hidden layers, following the convention in Section~\ref{sec:methodology}, write
\[
z^{(0)}=x,\qquad
s^{(k)}=W^{(k)}z^{(k-1)},\qquad
z^{(k)}=\operatorname{ReLU}(s^{(k)}),\quad k\in[K],
\]
and \(f(x)=W^{(K+1)}z^{(K)}\). For \(q_y=e_y-e_{\bar y}\), the objective is
\[
f(x)_y-f(x)_{\bar y}=c^\top z^{(K)},
\qquad c=(W^{(K+1)})^\top q_y.
\]

Starting from \([\ell_z^{(0)},u_z^{(0)}]=[\ell_x,u_x]\), interval bounds can be propagated by
\[
\begin{aligned}
\ell_s^{(k)}&=W_+^{(k)}\ell_z^{(k-1)}+W_-^{(k)}u_z^{(k-1)},\\
u_s^{(k)}&=W_+^{(k)}u_z^{(k-1)}+W_-^{(k)}\ell_z^{(k-1)},\\
\ell_z^{(k)}&=\max\{0,\ell_s^{(k)}\},\qquad
u_z^{(k)}=\max\{0,u_s^{(k)}\},
\end{aligned}
\]
where \(W_+=\max\{W,0\}\), \(W_-=\min\{W,0\}\), and the operations are componentwise. At every hidden layer, impose these boxes, the linear equation for \(s^{(k)}\), and the ReLU hull in~\eqref{convex-hull}. Convolutions are linear operators under this convention, so the same construction applies after indexing their input and output coordinates.

\paragraph{Diagonal and product lifts.}
Let \(\zeta^{(0)}=\xi\). A multilayer formulation can impose the valid square bounds
\[
(z_i^{(k)})^2\le\zeta_i^{(k)}
\le(\ell_{z,i}^{(k)}+u_{z,i}^{(k)})z_i^{(k)}
-\ell_{z,i}^{(k)}u_{z,i}^{(k)},\qquad k=1,\ldots,K.
\]

For the cross-layer, current-layer, and previous-layer couplings in Figure~\ref{fig:SOCP}, introduce
\[
\begin{aligned}
\eta_{ij}^{(k)}&\approx z_i^{(k-1)}z_j^{(k)}, &&(i,j)\in\mathcal E^{(k)},\\
\sigma_{jj'}^{(k)}&\approx z_j^{(k)}z_{j'}^{(k)}, &&(j,j')\in\mathcal S^{(k)},\ j<j',\\
\tau_{pq}^{(k)}&\approx z_p^{(k-1)}z_q^{(k-1)}, &&(p,q)\in\mathcal T^{(k)},\ p<q.
\end{aligned}
\]
Each product variable \(w\approx ab\) over \([\ell_a,u_a]\times[\ell_b,u_b]\) satisfies
\[
\begin{aligned}
w&\ge\ell_a b+\ell_b a-\ell_a\ell_b,&
w&\ge u_a b+u_b a-u_a u_b,\\
w&\le u_a b+\ell_b a-u_a\ell_b,&
w&\le\ell_a b+u_b a-\ell_a u_b.
\end{aligned}
\]
With shared diagonal surrogates \(d_a,d_b\) for \(a^2,b^2\), impose the local PSD constraint in SOC form:
\[
\begin{bmatrix}d_a&w\\w&d_b\end{bmatrix}\succeq0
\quad\Longleftrightarrow\quad
\left\|\begin{bmatrix}2w\\d_a-d_b\end{bmatrix}\right\|_2\le d_a+d_b.
\]
For previous-layer couplings \(\mathcal T^{(k)}\), the diagonal entries are \(\zeta_p^{(k-1)}\) and \(\zeta_q^{(k-1)}\), reducing to \(\xi_p\) and \(\xi_q\) at \(k=1\). Each diagonal variable is shared across all pairs and adjacent layers.

\paragraph{Residual identities.}
At each layer, ReLU complementarity gives \((z_j^{(k)})^2=z_j^{(k)}s_j^{(k)}\), yielding
\begin{equation}\label{eq:multilayer-zeta}
\zeta_j^{(k)}
=\sum_{i:(i,j)\in\mathcal E^{(k)}}W_{ji}^{(k)}\eta_{ij}^{(k)}
+r_j^{(k)},
\qquad r_j^{(k)}\in[L_j^{(r,k)},U_j^{(r,k)}].
\end{equation}
The residual collects the omitted terms
\(\sum_{i:(i,j)\notin\mathcal E^{(k)}}W_{ji}^{(k)}z_i^{(k-1)}z_j^{(k)}\),
bounded using Appendix~\ref{app:residual-bounds}. For one hidden layer,
\(z^{(0)}=x\) and \(W^{(1)}=W_1\), so~\eqref{eq:multilayer-zeta} reduces to~\eqref{eq:zeta-residual-main}.
With the layer-wise constraints above, the exact-lift argument in Appendix~\ref{app:proof-socp-upper-bound} establishes a valid upper bound on the output margin.

\section{Sparse Coupling Selection Details}
\label{app:sparse-coupling-selection}

\paragraph{Backward influence coefficients.}
Using the notation of Appendix~\ref{sec:socp-multilayer}, let \(q_y=e_y-e_{\bar y}\). The released evaluator defaults to \texttt{gamma\_backend=dual}, which computes CROWN-style influence coefficients~\citep{zhang2018efficient}; \texttt{dual\_WK} selects the implemented Wong-Kolter-style recursion~\citep{wong2018provable}. This choice affects coupling selection, independently of the model's training method. For an unstable ReLU, define
\[
\alpha_j^{(k)}=\frac{u_{s,j}^{(k)}}{u_{s,j}^{(k)}-\ell_{s,j}^{(k)}},
\qquad
\beta_j^{(k)}=\mathbf 1\{u_{s,j}^{(k)}> -\ell_{s,j}^{(k)}\}.
\]
For the CROWN-style recursion, initialize \(\lambda^{(K+1)}=q_y\). For \(k=K,\ldots,1\), set
\[
\widehat\lambda^{(k)}=(W^{(k+1)})^\top\lambda^{(k+1)},
\qquad
\gamma_j^{(k)}=\lambda_j^{(k)}=
\begin{cases}
0, & u_{s,j}^{(k)}\le0,\\
\widehat\lambda_j^{(k)}, & \ell_{s,j}^{(k)}\ge0,\ u_{s,j}^{(k)}>0,\\
\alpha_j^{(k)}\widehat\lambda_j^{(k)}, & \ell_{s,j}^{(k)}<0<u_{s,j}^{(k)},\ \widehat\lambda_j^{(k)}\ge0,\\
\beta_j^{(k)}\widehat\lambda_j^{(k)}, & \ell_{s,j}^{(k)}<0<u_{s,j}^{(k)},\ \widehat\lambda_j^{(k)}<0.
\end{cases}
\]
For \texttt{dual\_WK}, initialize \(\nu^{(K+1)}=-q_y\) and propagate
\(\widehat\nu^{(k)}=(W^{(k+1)})^\top\nu^{(k+1)}\). Stable units use the same inactive/active rules, while the implemented unstable-unit step is
\[
\gamma_j^{(k)}=\nu_j^{(k)}
=\alpha_j^{(k)}[\widehat\nu_j^{(k)}]_+
+[\widehat\nu_j^{(k)}]_-,
\qquad [a]_+=\max\{a,0\},\quad [a]_-=\min\{a,0\}.
\]
These expressions state the unregularized recursions; the code adds \(10^{-12}\) to the slope denominators. Convolutional maps use the corresponding adjoint convolution.

\paragraph{Candidate pools and pair scores.}
Rank current-layer neurons by
\[
\rho_j^{(k)}=
\begin{cases}
|\gamma_j^{(k)}|(u_{s,j}^{(k)}-\ell_{s,j}^{(k)}),
&\ell_{s,j}^{(k)}<0<u_{s,j}^{(k)},\\
-\infty,&\text{otherwise}.
\end{cases}
\]
The \(\min\{N_z,n_k\}\) largest scores form \(\mathcal R^{(k)}\). This prioritizes unstable units; if fewer than \(N_z\) are unstable, the released top-\(N_z\) routine also includes stable units. For the preceding layer, define
\(\Delta z_i^{(k-1)}=u_{z,i}^{(k-1)}-\ell_{z,i}^{(k-1)}\), and let \(\mathcal C^{(k-1)}\) contain the \(\min\{N_p,n_{k-1}\}\) widest intervals. With \(i,p,q\in\mathcal C^{(k-1)}\) and \(j,j'\in\mathcal R^{(k)}\), score the pairs by
\[
\begin{aligned}
r_{\mathcal E}^{(k)}(i,j)
&=|\gamma_j^{(k)}|\,|W_{ji}^{(k)}|\,\Delta z_i^{(k-1)}u_{z,j}^{(k)},\\
r_{\mathcal S}^{(k)}(j,j')
&=|\gamma_j^{(k)}\gamma_{j'}^{(k)}|\,u_{z,j}^{(k)}u_{z,j'}^{(k)},\qquad j<j',\\
A_p^{(k)}&=\sum_{j\in\mathcal R^{(k)}}|\gamma_j^{(k)}|\,|W_{jp}^{(k)}|,\\
r_{\mathcal T}^{(k)}(p,q)
&=A_p^{(k)}A_q^{(k)}\Delta z_p^{(k-1)}\Delta z_q^{(k-1)},\qquad p<q.
\end{aligned}
\]
At each layer, retain the highest positive, finite-scoring pairs, up to \(B_E\), \(B_S\), and \(B_T\), respectively. Previous-layer couplings use \(z^{(k-1)}\): the inputs \(x\) at \(k=1\) and hidden activations at \(k\ge2\). Budgets remain fixed within a run, while scores and selected sets are recomputed for each input and target. Record the selected indices with the software version to reproduce choices among equal scores.

\section{Computational Complexity and Resource-Limited Verification}
\label{appendix:time_complexity}
\label{app:resource-limited-verification}
This section relates the resource-limited results in Table~\ref{tab:robust-trained-ab-crown} to the formulation and solver costs in Remark~\ref{rem:complexity}.

\paragraph{Observed memory-limited evaluations.}
For the CROWN-IBP and DeepPoly MNIST models trained at \(\varepsilon_{\mathrm{train}}=0.1\), evaluation at \(\varepsilon_{\mathrm{verify}}=0.15\) exhausts the available memory during the \(\alpha,\beta\)-CROWN runs. The process exits are caused by memory constraints rather than expiration of the 120-second per-property timeout. The evaluations use the configuration in Appendix~\ref{app:experimental-optimization}; smaller-batch retries also fail to complete. Consequently, N/A denotes unavailable certified robust accuracy, not a zero value or a determination of non-robustness. SOCP-cert completes the corresponding 300-image evaluations, certifying 8.00\% and 7.33\%, respectively.

\paragraph{Fixed SOCP formulation.}
For \(K\) hidden layers of widths \(n_1,\ldots,n_K\), with \(n_0=d\), define
\[
N=d+\sum_{k=1}^{K}n_k,\qquad
Q=\sum_{k=1}^{K}\bigl(|\mathcal E^{(k)}|+|\mathcal S^{(k)}|+|\mathcal T^{(k)}|\bigr).
\]
The base variables and square lifts number \(O(N)\). Each selected pair adds one product variable, four McCormick inequalities, and a constant-dimensional SOC; square epigraphs also use constant-dimensional SOCs. Thus the formulation has \(O(N+Q)\) variables, constraints, and cones, compared with \(O(N^2)\) entries in a dense SDP lift. If \(A\) counts nonzero coefficients in the expanded network operators, sparse conic data require \(O(A+N+Q)\) storage. For convolutions, \(A\) counts repeated weight uses, not only trainable parameters.

\paragraph{Construction and solution costs.}
With sparse operators available, interval propagation costs \(O(A+N)\) per input. For \(p_k=|\mathcal C^{(k-1)}|\) and \(q_k=|\mathcal R^{(k)}|\) in Appendix~\ref{app:sparse-coupling-selection}, define
\[
P_{\mathrm{cand}}=\sum_{k=1}^{K}
\left(p_kq_k+\binom{q_k}{2}+\binom{p_k}{2}\right).
\]
Backward influence, comparison sorting, residual bounds, and assembly give the per-target bound
\[
T_{\mathrm{form}}=O\!\left(A+N\log(N+1)
+P_{\mathrm{cand}}\log(P_{\mathrm{cand}}+1)+Q\right).
\]
For Newton-system dimension \(M=O(N+Q)\) and \(I_{\mathrm{SOCP}}\) interior-point iterations, conventional dense factorization gives
\[
T_{\mathrm{target}}=O(T_{\mathrm{form}}+I_{\mathrm{SOCP}}M^3),\qquad
S_{\mathrm{SOCP}}=O(A+N+Q+P_{\mathrm{cand}}+M^2).
\]
Under strict feasibility and suitable primal-dual initialization, short-step path following requires
\(I_{\mathrm{SOCP}}=O(\sqrt{N+Q}\log(g_0/\delta))\)
for initial duality gap \(g_0\) and tolerance \(0<\delta<g_0\)~\citep{alizadeh2003second,ben2001lectures}. This yields a polynomial arithmetic-operation bound for a fixed relaxation, excluding initialization. Sparse factorization depends on fill-in; SCS has different iteration costs. Dense operator conversion adds preprocessing and \(O(\sum_{k=1}^{K+1}n_{k-1}n_k)\) storage, with \(n_{K+1}=L\).

\paragraph{Branch-and-bound time and memory.}
\(\alpha,\beta\)-CROWN combines bound propagation with batched branching and retains unresolved subdomains~\citep[Appendix~B]{wang2021beta}. If root processing costs \(T_{\mathrm{root}}\), at most \(D\) subdomains are processed, and each costs at most \(T_{\mathrm{node}}\), then
\[
T_{\mathrm{BaB}}=O(T_{\mathrm{root}}+D T_{\mathrm{node}}).
\]
An uncapped binary search over \(U\) initially unstable ReLUs can have \(D=O(2^U)\); this is a search-tree bound, not the cost of one bound-propagation pass. Peak memory depends instead on retained domains and simultaneous computation. With at most \(R\) stored domains, per-domain storage bounded by \(s_d\), batch workspace \(W(b)\), and model/root-processing storage \(S_0\), a conservative bound is
\[
S_{\mathrm{BaB}}=O(S_0+R s_d+W(b)).
\]
Bounds, split histories, and relaxation parameters contribute to this state; CPU and GPU allocations each face their own capacity limits. Reducing batch size \(b\) lowers concurrent work but does not eliminate root-processing or retained-domain storage. Neither a timeout nor a domain-count cap is a bound on all memory allocations. For an uncapped search that retains its frontier, \(R\le D=O(2^U)\) gives the conservative worst-case bound \(O(S_0+2^U s_d+W(b))\). This allows exponential frontier growth; traversal policies that retain fewer domains can use substantially less memory.

\paragraph{Why limited memory can favor SOCP.}
For difficult properties with many unresolved subdomains, SOCP avoids search-frontier growth by solving a fixed relaxation whose size is controlled by the coupling budgets. Candidate-pool limits separately control selection work. This provides a practical advantage in the reported memory-limited runs: it completes an evaluation with useful certificates while the configured complete-search runs terminate. The trade-off is incompleteness; a fixed SOCP may leave robust inputs uncertified. Factorization and operator storage can still be substantial, so these results support a configuration-dependent advantage in evaluation completion, without establishing universal memory or runtime superiority. More memory or different settings may allow \(\alpha,\beta\)-CROWN to complete.


%% file: references.bib
@article{kim2001second,
  title={Second order cone programming relaxation of nonconvex quadratic optimization problems},
  author={Kim, Sunyonga and Kojima, Masakazu},
  journal={Optimization methods and software},
  volume={15},
  number={3-4},
  pages={201--224},
  year={2001},
  publisher={Taylor \& Francis}
}

@article{raghunathan2018semidefinite,
  title={Semidefinite relaxations for certifying robustness to adversarial examples},
  author={Raghunathan, Aditi and Steinhardt, Jacob and Liang, Percy S},
  journal={Advances in neural information processing systems},
  volume={31},
  year={2018}
}

@inproceedings{wong2018provable,
  title={Provable Defenses against Adversarial Examples via the Convex Outer Adversarial Polytope},
  author={Wong, Eric and Kolter, Zico},
  booktitle={International Conference on Machine Learning},
  pages={5286--5295},
  year={2018}
}

@article{wang2021beta,
  title={Beta-crown: Efficient bound propagation with per-neuron split constraints for neural network robustness verification},
  author={Wang, Shiqi and Zhang, Huan and Xu, Kaidi and Lin, Xue and Jana, Suman and Hsieh, Cho-Jui and Kolter, J Zico},
  journal={Advances in Neural Information Processing Systems},
  volume={34},
  year={2021}
}

@inproceedings{lofberg2004yalmip,
  title={YALMIP: A toolbox for modeling and optimization in MATLAB},
  author={Lofberg, Johan},
  booktitle={2004 IEEE international conference on robotics and automation (IEEE Cat. No. 04CH37508)},
  pages={284--289},
  year={2004},
  organization={IEEE}
}

@article{maleki2026cascading,
  title={Cascading Robustness Verification: Toward Efficient Model-Agnostic Certification},
  author={Maleki, Mohammadreza and Sidibomma, Rushendra and Adibi, Arman and Samavi, Reza},
  journal={arXiv preprint arXiv:2602.04236},
  year={2026}
}

@article{chiu2025sdp,
  title={SDP-CROWN: Efficient Bound Propagation for Neural Network Verification with Tightness of Semidefinite Programming},
  author={Chiu, Hong-Ming and Chen, Hao and Zhang, Huan and Zhang, Richard Y},
  journal={arXiv preprint arXiv:2506.06665},
  year={2025}
}

@article{mccormick1976computability,
  title={Computability of Global Solutions to Factorable Nonconvex Programs: Part I—Convex Underestimating Problems},
  author={McCormick, G. P.},
  journal={Mathematical Programming},
  volume={10},
  pages={147--175},
  year={1976}
}

@book{ben2001lectures,
  title={Lectures on modern convex optimization: analysis, algorithms, and engineering applications},
  author={Ben-Tal, Aharon and Nemirovski, Arkadi},
  year={2001},
  publisher={SIAM}
}

@article{alizadeh2003second,
  title={Second-order cone programming},
  author={Alizadeh, Farid and Goldfarb, Donald},
  journal={Mathematical programming},
  volume={95},
  number={1},
  pages={3--51},
  year={2003}
}

@article{pulina2012challenging,
  title={Challenging SMT solvers to verify neural networks},
  author={Pulina, Luca and Tacchella, Armando},
  journal={Ai Communications},
  volume={25},
  number={2},
  pages={117--135},
  year={2012},
  publisher={IOS Press}
}

@inproceedings{katz2017reluplex,
  title={Reluplex: An efficient SMT solver for verifying deep neural networks},
  author={Katz, Guy and Barrett, Clark and Dill, David L and Julian, Kyle and Kochenderfer, Mykel J},
  booktitle={International conference on computer aided verification},
  pages={97--117},
  year={2017},
  organization={Springer}
}

@article{fazlyab2020safety,
  title={Safety verification and robustness analysis of neural networks via quadratic constraints and semidefinite programming},
  author={Fazlyab, Mahyar and Morari, Manfred and Pappas, George J},
  journal={IEEE Transactions on Automatic Control},
  year={2020},
  publisher={IEEE}
}

@inproceedings{wong2018scaling,
  title={Scaling provable adversarial defenses},
  author={Wong, Eric and Schmidt, Frank and Metzen, Jan Hendrik and Kolter, J Zico},
  booktitle={Advances in Neural Information Processing Systems},
  pages={8400--8409},
  year={2018}
}

@article{zhang2022general,
  title={General cutting planes for bound-propagation-based neural network verification},
  author={Zhang, Huan and Wang, Shiqi and Xu, Kaidi and Li, Linyi and Li, Bo and Jana, Suman and Hsieh, Cho-Jui and Kolter, J Zico},
  journal={Advances in neural information processing systems},
  volume={35},
  pages={1656--1670},
  year={2022}
}

@inproceedings{dathathri2020enabling,
 author = {Dathathri, Sumanth and Dvijotham, Krishnamurthy and Kurakin, Alexey and Raghunathan, Aditi and Uesato, Jonathan and Bunel, Rudy R and Shankar, Shreya and Steinhardt, Jacob and Goodfellow, Ian and Liang, Percy S and Kohli, Pushmeet},
 booktitle = {Advances in Neural Information Processing Systems},
 editor = {H. Larochelle and M. Ranzato and R. Hadsell and M. F. Balcan and H. Lin},
 pages = {5318--5331},
 title = {Enabling certification of verification-agnostic networks via memory-efficient semidefinite programming},
 volume = {33},
 year = {2020}
}

@inproceedings{lan2022tight,
  title={Tight neural network verification via semidefinite relaxations and linear reformulations},
  author={Lan, Jianglin and Zheng, Yang and Lomuscio, Alessio},
  booktitle={Proceedings of the AAAI Conference on Artificial Intelligence},
  volume={36},
  number={7},
  pages={7272--7280},
  year={2022}
}

@article{xu2020fast,
  title={Fast and complete: Enabling complete neural network verification with rapid and massively parallel incomplete verifiers},
  author={Xu, Kaidi and Zhang, Huan and Wang, Shiqi and Wang, Yihan and Jana, Suman and Lin, Xue and Hsieh, Cho-Jui},
  journal={arXiv preprint arXiv:2011.13824},
  year={2020}
}

@inproceedings{li2023sok,
  title={Sok: Certified robustness for deep neural networks},
  author={Li, Linyi and Xie, Tao and Li, Bo},
  booktitle={2023 IEEE symposium on security and privacy (SP)},
  pages={1289--1310},
  year={2023},
  organization={IEEE}
}

@inproceedings{
    madry2017towards,
    title={Towards Deep Learning Models Resistant to Adversarial Attacks},
    author={Aleksander Madry and Aleksandar Makelov and Ludwig Schmidt and Dimitris Tsipras and Adrian Vladu},
    booktitle={International Conference on Learning Representations},
    year={2018},
}

@article{zhang2018efficient,
  title={Efficient neural network robustness certification with general activation functions},
  author={Zhang, Huan and Weng, Tsui-Wei and Chen, Pin-Yu and Hsieh, Cho-Jui and Daniel, Luca},
  journal={Advances in neural information processing systems},
  volume={31},
  year={2018}
}

@article{huang2017adversarial,
  title={Adversarial attacks on neural network policies},
  author={Huang, Sandy and Papernot, Nicolas and Goodfellow, Ian and Duan, Yan and Abbeel, Pieter},
  journal={arXiv preprint arXiv:1702.02284},
  year={2017}
}

@article{vandenberghe1996semidefinite,
  title={Semidefinite programming},
  author={Vandenberghe, Lieven and Boyd, Stephen},
  journal={SIAM review},
  volume={38},
  number={1},
  pages={49--95},
  year={1996},
  publisher={SIAM}
}

@inproceedings{gowal2018ibp,
  title={On the Effectiveness of Interval Bound Propagation for Training Verifiably Robust Models},
  author={Gowal, Sven and Qin, Chongli and Uesato, Jonathan and Mann, Timothy and Kohli, Pushmeet},
  booktitle={arXiv preprint arXiv:1810.12715},
  year={2018}
}

@inproceedings{zhang2020crownibp,
  title={Towards Stable and Efficient Training of Verifiably Robust Neural Networks},
  author={Zhang, Huan and Chen, Hongge and Xiao, Chaowei and Gowal, Sven and Stanforth, Robert and Li, Bo and Boning, Duane and Hsieh, Cho-Jui},
  booktitle={International Conference on Learning Representations},
  year={2020}
}

@article{mao2024ctbench,
  title={{CTBench}: A Library and Benchmark for Certified Training},
  author={Mao, Yuhao and Balauca, Stefan and Vechev, Martin},
  journal={arXiv preprint arXiv:2406.04848},
  year={2024}
}

@inproceedings{singh2019deeppoly,
  title={An Abstract Domain for Certifying Neural Networks},
  author={Singh, Gagandeep and Gehr, Timon and P{\"u}schel, Markus and Vechev, Martin},
  booktitle={Proceedings of the ACM on Programming Languages (POPL)},
  volume={3},
  number={POPL},
  pages={1--30},
  year={2019}
}

@article{sadig2026cear,
  title={CEAR: Certified Ensemble Adversarial Robustness in DNNs},
  author={Sadig, Daniel and Maleki, Mohammadreza and Karimi, Hamed and Samavi, Reza},
  journal={arXiv preprint arXiv:2606.01437},
  year={2026}
}

@inproceedings{tjeng2019evaluating,
  title={Evaluating Robustness of Neural Networks with Mixed Integer Programming},
  author={Tjeng, Vincent and Xiao, Kai and Tedrake, Russ},
  booktitle={International Conference on Learning Representations},
  year={2019}
}

@misc{LeCun_Burges_Cortes_2010,
  author       = {Yann LeCun and Corinna Cortes and Christopher J. C. Burges},
  title        = {The MNIST Database of Handwritten Digits},
  year         = {1998},
  howpublished = {\url{http://yann.lecun.com/exdb/mnist/}}
}

@misc{Krizhevsky_2012,
  author       = {Alex Krizhevsky},
  title        = {Learning Multiple Layers of Features from Tiny Images},
  year         = {2009},
  howpublished = {Technical Report, University of Toronto},
  note         = {CIFAR-10 dataset},
  url          = {https://www.cs.toronto.edu/~kriz/learning-features-2009-TR.pdf}
}
